\documentclass[lettersize,journal]{IEEEtran}
\usepackage{amsmath,amsfonts}
\usepackage{algorithmic}
\usepackage{algorithm}
\usepackage{array}
\usepackage[caption=false,font=normalsize,labelfont=sf,textfont=sf]{subfig}
\usepackage{textcomp}
\usepackage{stfloats}
\usepackage{url}
\usepackage{verbatim}
\usepackage{graphicx}
\usepackage{cite}
\usepackage{tikz}
\usepackage{comment}
\usepackage{color}
\usepackage{caption}
\usepackage{epsfig}
\usepackage{url}
\usepackage{comment}
\usepackage{color}
\usepackage{enumitem}
\usepackage{booktabs}
\usepackage{graphicx}
\usepackage{bm}
\usepackage{multirow}
\usepackage{wrapfig}
\usepackage{xcolor}
\usepackage{color, colortbl}

\renewcommand{\vec}[1]{\boldsymbol{#1}}    

\def\BibTeX{{\rm B\kern-.05em{\sc i\kern-.025em b}\kern-.08em
    T\kern-.1667em\lower.7ex\hbox{E}\kern-.125emX}}

\begin{document}

\title{Towards Real-World Video Denosing: A Practical Video Denosing Dataset and Network}

\author{
Xiaogang Xu$\dag$, Yitong Yu$\dag$, Nianjuan Jiang, Bei Yu,~\IEEEmembership{Senior Member,~IEEE}, \\Jiangbo Lu,~\IEEEmembership{Senior Member,~IEEE}, and Jiaya Jia,~\IEEEmembership{Fellow,~IEEE}
\IEEEcompsocitemizethanks{
\IEEEcompsocthanksitem $\dag$ indicates equal contribution.
\IEEEcompsocthanksitem X.~Xu are with the Department of Computer Science and Engineering, The Chinese University of Hong Kong. He is also an affiliated researcher of SmartMore. E-mail: xgxu@cse.cuhk.edu.hk.
\IEEEcompsocthanksitem Y.~Yu is a researcher in Sensetime Research and Tetras AI. E-mail: yuyitong@tetras.ai.
\IEEEcompsocthanksitem N.~Jiang, and J.~Lu are with SmartMore.
E-mail: \{yitong.yu, nianjuan.jiang, jiangbo.lu\}@smartmore.com.
\IEEEcompsocthanksitem B.~Yu, and J.~Jia are with the Department of Computer Science and Engineering, The Chinese University of Hong Kong.
E-mail: \{byu, leojia\}@cse.cuhk.edu.hk.
\IEEEcompsocthanksitem This work is supported by Shenzhen Science and Technology Program (KQTD20210811090149095).
}}

\markboth{submission to IEEE Transactions on Image Processing}
{Shell \MakeLowercase{\textit{et al.}}: A Sample Article Using IEEEtran.cls for IEEE Journals}


\maketitle

\begin{abstract}
To facilitate video denoising research, we construct a compelling dataset, namely, ``Practical Video Denoising Dataset" (PVDD), containing 200 noisy-clean dynamic video pairs in both sRGB and RAW format.
Compared with existing datasets consisting of limited motion information,
PVDD covers dynamic scenes with varying and natural motion. Different from datasets using primarily Gaussian or Poisson distributions to synthesize noise in the sRGB domain, PVDD synthesizes realistic noise from the RAW domain with a physically meaningful sensor noise model followed by ISP processing.
Moreover, we also propose a new video denoising framework, called Recurrent Video Denoising Transformer (RVDT), which can achieve SOTA performance on PVDD and other current video denoising benchmarks.
RVDT consists of both spatial and temporal transformer blocks to conduct denoising with long-range operations on the spatial dimension and long-term propagation on the temporal dimension.
Especially, RVDT exploits the attention mechanism to implement the bi-directional feature propagation with both implicit and explicit temporal modeling. 
Extensive experiments demonstrate that 1) models trained on PVDD achieve superior denoising performance on many challenging real-world videos than on models trained on other existing datasets; 2) trained on the same dataset, our proposed RVDT can have better denoising performance than other types of networks.
\end{abstract}

\begin{IEEEkeywords}
Video Denoising Dataset, Dynamic Scenes, Realistic Noises, Recurrent Video Denoising Transformer
\end{IEEEkeywords}

\section{Introduction}

Signal degradation caused by the sensor noise is common when capturing videos, resulting in poor visual quality.
Therefore, video denoising is a task with significant importance.
Deep-learning-based denoising techniques have emerged in recent years~\cite{tassano2020fastdvdnet,tassano2019dvdnet,maggioni2021efficient,claus2019videnn,yue2020supervised}.
These methods learn the mapping from the noisy image/video domain to the clean image/video domain, while the corresponding performance heavily depends on the characteristic of the training data.

Although practical image denoising datasets have been introduced to the community, they are still rather limited regarding the diversity in motion. Directly employing them for the video denoising task generally results in the lack of variety in temporal space~\cite{yue2020supervised} and thus cannot provide sufficient
training features for challenging video processing tasks. More practical and capable video denoising datasets, which fall short currently, are in critical demand.

\begin{figure}[t]
    \centering
    \includegraphics[width=1.0\linewidth]{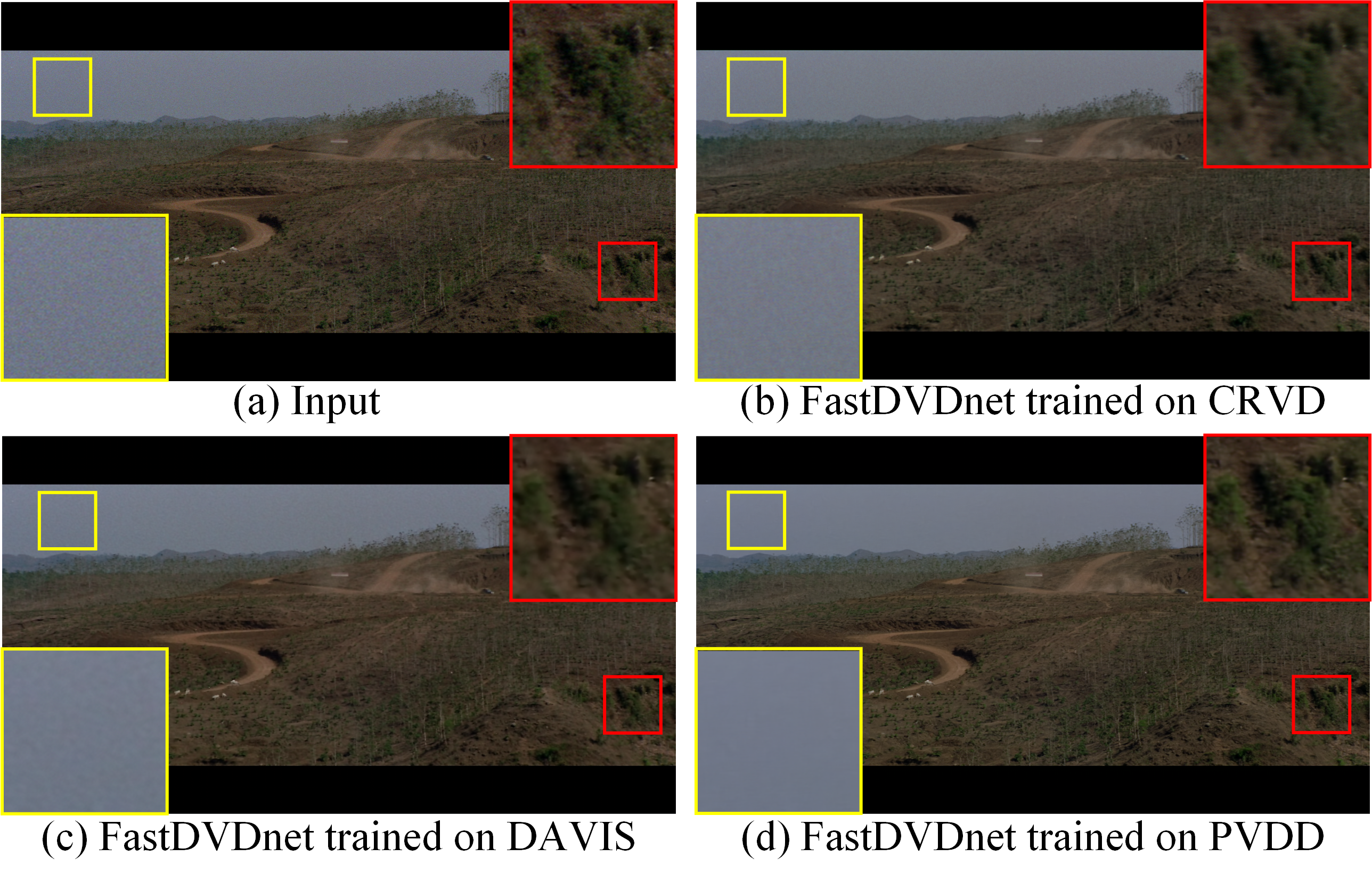}
    \caption{
    Models trained on our PVDD yield superior performance on real-world noisy videos. (a) An example frame of a challenging video with the complicated noise pattern that never appears in the training set. 
    (b)-(d) show output of FastDVDnet~\cite{tassano2020fastdvdnet} trained on different datasets.
    The model trained on our PVDD achieves the best denoising performance (yellow boxes in (a)-(d)) and preserves finer details without introducing blur and noise (red boxes in (a)-(d)).
    }
    \label{fig:teaser}
\end{figure}

\begin{figure}[t]
    \centering
    \includegraphics[width=1.0\linewidth]{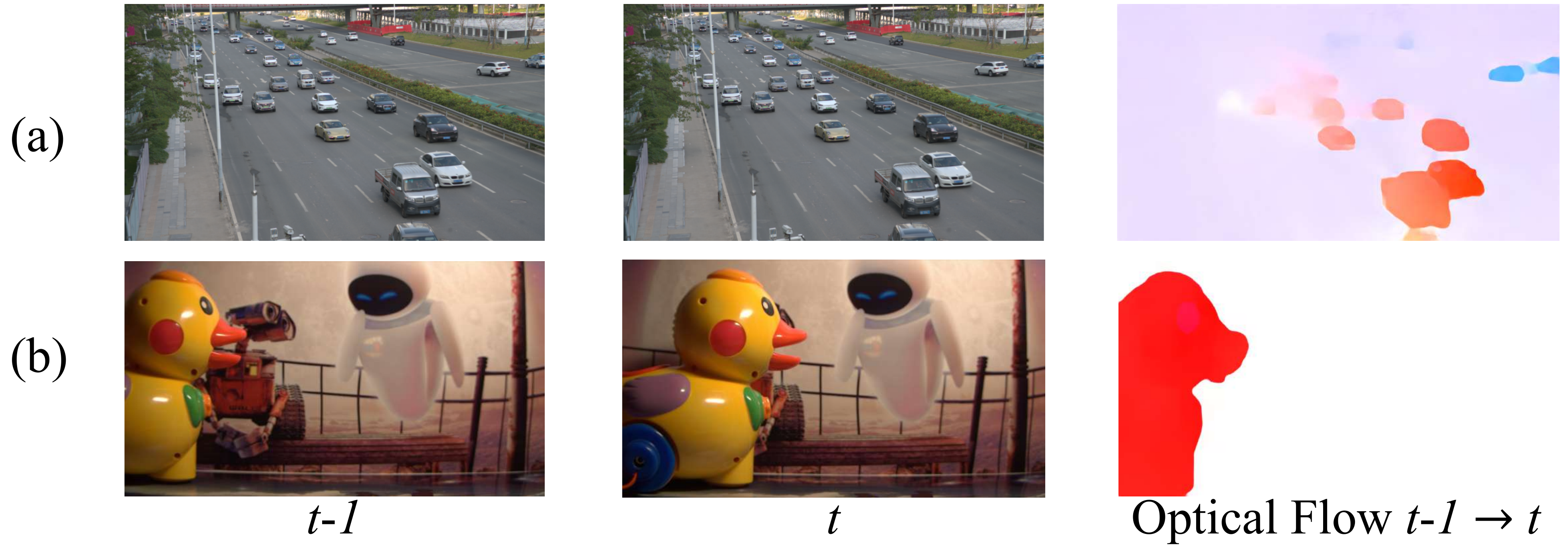}
    \caption{
    Example frames with their optical flows from PVDD and CRVD. (a) Frames from PVDD are captured in video mode, containing complex natural motion. (b) Frames from CRVD are captured at discrete time instances and only contain simple and overly large frame-to-frame foreground motion.
    }
    \label{fig:example1}
\end{figure}

There are two categories of video denoising datasets currently. The first kind collects video pairs. It creates clean video frames by averaging neighboring noisy frames~\cite{yue2020supervised}. 
It requires video frames to be either static or captured at discrete time instances, as shown in Fig.~\ref{fig:example1}(b), and cannot provide training data with complex natural motion.
The second group, e.g., DAVIS~\cite{perazzi2016benchmark}, creates noisy-clean pairs with added synthetic noise. It first collects clean videos and introduces noise (e.g., Gaussian noise) in the sRGB domain. The added noise may not be the same or even similar to that produced by cameras, considering complex distributions of real noise~\cite{nam2016holistic,zhou2020awgn}.

In addition, as explained in~\cite{zhou2020awgn,yue2020supervised,abdelhamed2018highsidd}, shot noise in RAW format can be modeled by Poisson distributions and read noise follows Gaussian distributions. Recent work~\cite{zhang2021rethinking} further shows that the denoising networks trained with the Poisson-Gaussian noise distribution assumption achieve comparable results with networks trained on data degraded by real noise in the RAW domain. However, no existing datasets take this important fact into consideration yet when collecting data.

\begin{figure}[t]
    \centering
    \includegraphics[width=1.0\linewidth]{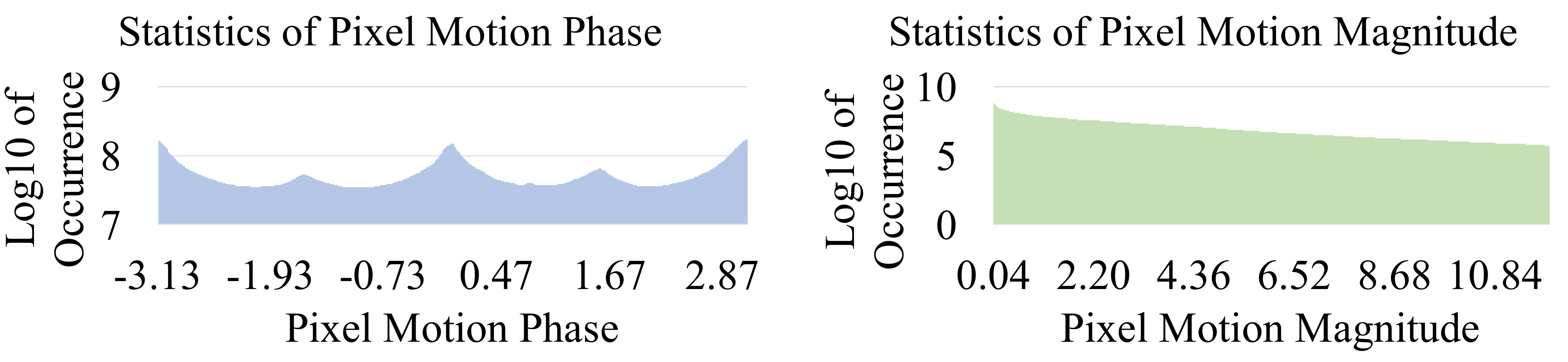}
    \caption{
    Motion statistics in PVDD in terms of motion phase and magnitude, exhibiting the richness of natural motion. The per-pixel flow in every frame of PVDD is computed via a representative OpenCV algorithm~\cite{farneback2003two}.
    }
    \label{fig:example2}
\end{figure}

\begin{figure}[t]
    \centering
    \includegraphics[width=1.0\linewidth]{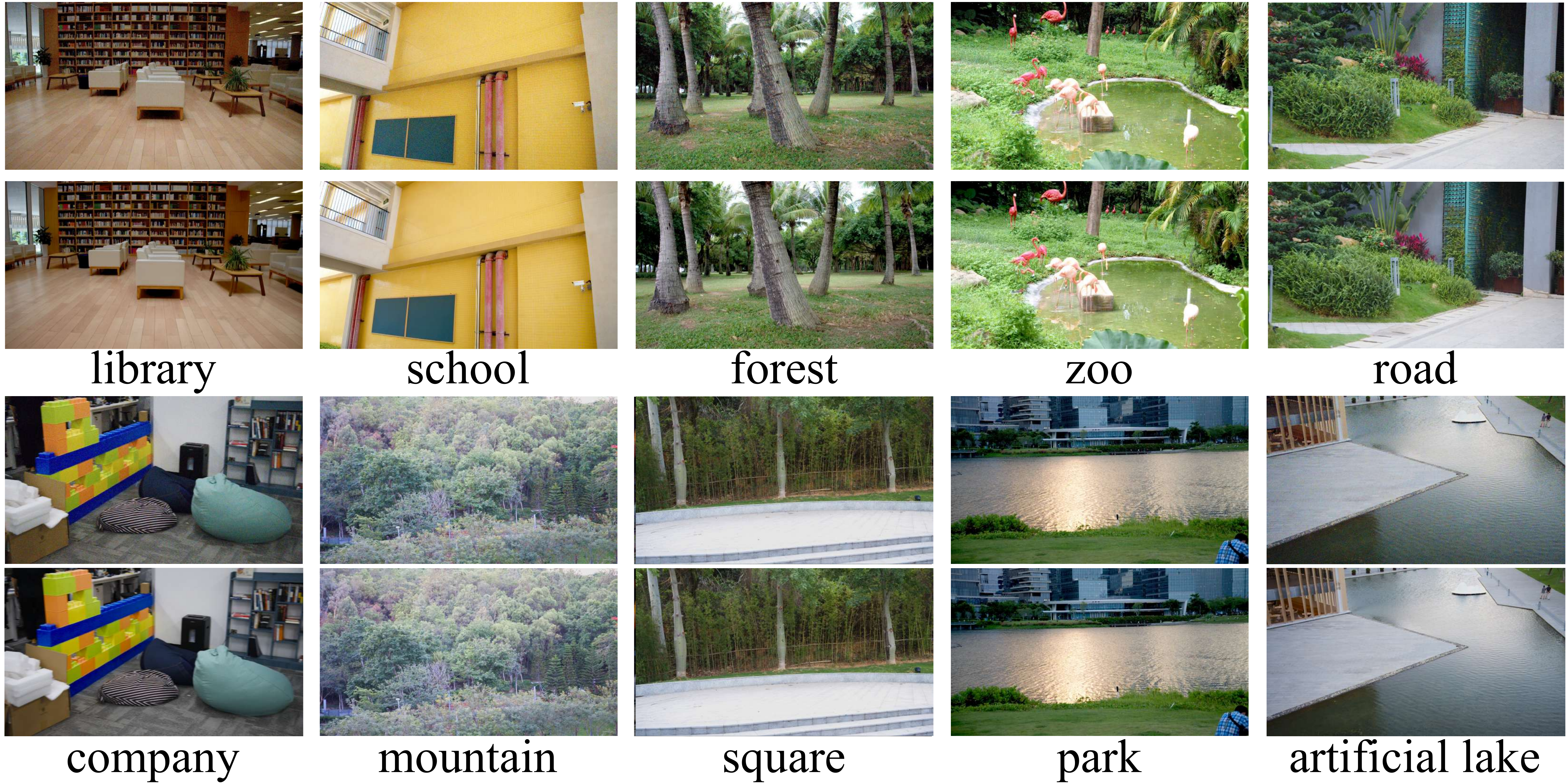}
    \caption{
    Visual examples to demonstrate the diversity of our dataset, e.g., regarding different indoor/outdoor scenes.
    }
    \label{fig:example-diver}
\end{figure}

In this paper, we construct a new powerful dataset, ``Practical Video Denoising Dataset" (PVDD), by collecting clean videos with complicated natural motion, followed by the corruption of realistic noise in RAW format to form RAW video pairs. The final sRGB video pairs are produced with a calibrated ISP processing pipeline. It contains 200 noisy-clean RAW/sRGB video pairs with rich dynamics, as shown in Figs.~\ref{fig:example1}(a) and~\ref{fig:example2}, which greatly diversify temporal local motion when used in system training. The clips also cover a variety of scenes, as illustrated in Fig.~\ref{fig:example-diver}, for better generalization for all networks. Table~\ref{data_cmp} provides a simple high-level comparison of PVDD with existing datasets -- the advantage is obvious.

In addition to the ISP modules utilized in previous work~\cite{abdelhamed2018highsidd,karaimer_brown_ECCV_2016}, we incorporate a color temperature module~\cite{robertson1968computation} and a filmic tone-mapper module~\cite{reinhard2002photographic} to better approximate the camera ISP pipeline (Fig.~\ref{fig:sRGB}(a)). 
Details will be presented in Sec.~\ref{sec:synthesis}. 
The resulting synthetic noises are realistic since ISP modules are calibrated, and noises added in the RAW domain are in accord with a realistic noise model.
PVDD can help the denoising for videos as shown in Fig.~\ref{fig:teaser}.

Besides the PVDD, we propose a novel video denosing framework in this paper, so-called recurrent Video Denoising Transformer (RVDT).
RVDT consists of spatial and bi-directional temporal transformer blocks. The spatial transformer blocks are concatenated with convolutional blocks to perform spatial-varying long-range and short-range operations.
The bi-directional temporal transformer blocks are built with both forward and backward blocks, where novel attention mechanisms are designed to achieve feature propagation from past and future frames. 
Further, long-range and short-range dependencies are simultaneously achieved in the transformers' feed-forward modules, enhancing the corresponding representation capacity.
Experiments are conducted on different datasets to show the superiority of the proposed RVDT to existing video denoising frameworks in both sRGB and RAW domains.

\textit{Our dataset and code will be made publicly available after publication as soon as possible}. In conclusion, our contribution is threefold.
\begin{itemize}
\setlength{\itemsep}{0pt}
\setlength{\parsep}{0pt}
\setlength{\parskip}{0pt}
    \item We propose an information-rich dataset, named PVDD, for video denoising in RAW and sRGB domains. It incorporates complex natural motion and realistic noise.
    \item Our PVDD serves as a practical benchmark dataset to evaluate video denoising networks and has the potential to further promote research along this line. We conduct extensive experiments with representative deep learning architectures including both quantitative and qualitative evaluation. They manifest the superiority of the proposed dataset and the new degradation model.
    \item We propose a novel video denoising framework RVDT with newly designed bi-directional transformer blocks.
    Extensive experiments are conducted on PVDD and existing video denoising benchmarks, demonstrating the SOTA performance of RVDT.
\end{itemize}

\section{Related Work}

\begin{table}[t]
    \centering
    \huge
\caption{Comparison between PVDD and other representative datasets.} 
    \label{data_cmp}
    \resizebox{1.0\linewidth}{!}{
    \begin{tabular}{c|p{3.5cm}<{\centering}p{3.5cm}<{\centering}p{3.9cm}<{\centering}p{3.5cm}<{\centering}p{3.5cm}<{\centering}p{3.5cm}<{\centering}}
        \toprule
        Dataset & Realistic Noise&Real Motions& Number of Scene & Resolution &sRGB & RAW \\
        \midrule
        CRVD~\cite{yue2020supervised}    &$\surd$& $\times$ & 11 & 1080P &$\times$&$\surd$  \\
        DAVIS~\cite{perazzi2016benchmark}  &$\times$& $\surd$& 90&480P&$\surd$&$\times$ \\\hline
        Ours & $\surd$ & $\surd$ & 200&1080P&$\surd$&$\surd$ \\
        \bottomrule
    \end{tabular}}
\end{table}

\subsection{Video Denoising Datasets}
Unlike image denoising datasets~\cite{nam2016holistic,yue2019high,brooks2019unprocessing,abdelhamed2018highsidd,anaya2018renoir,plotz2017benchmarkingdnd,chen2018learning,chen2019seeing}, video denoising datasets are far less, but still can be categorized into two kinds. 
The first kind collects video pairs by utilizing long exposure for clean videos and short exposure for noisy videos, e.g., SMID~\cite{chen2019seeing}, or creating clean video frames by averaging static neighboring noisy frames, e.g., CRVD~\cite{yue2020supervised}. 
These datasets contain only static or sparse frames captured in the temporal dimension.
The second kind creates noisy-clean pairs by noise synthesis, e.g., DAVIS~\cite{perazzi2016benchmark}. It adds noise (e.g., Gaussian noise) to clean videos.

\begin{figure*}[t]
    \centering
    \includegraphics[width=1.0\linewidth]{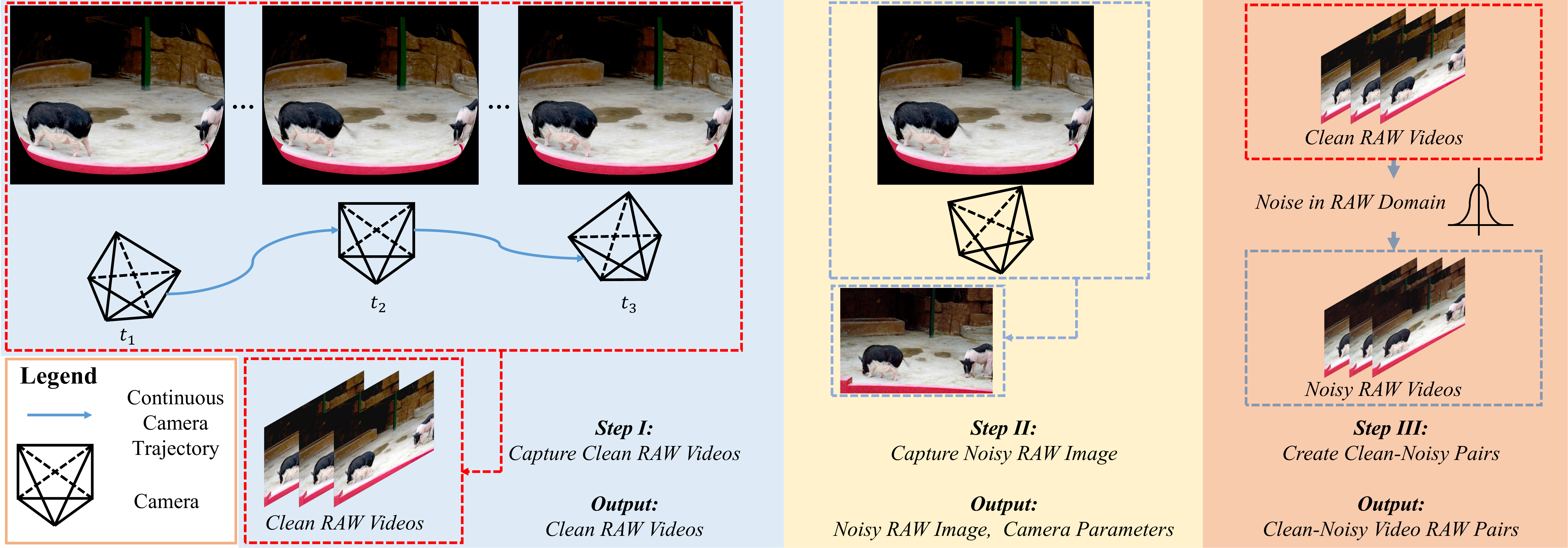}
    \vspace{-0.15in}
    \caption{
    Overview of the proposed pipeline to create dynamic noisy-clean video pairs in RAW format.
    }
    \vspace{-0.2in}
    \label{fig:framework}
\end{figure*}

\subsection{Video Denoising Methods}
Most video denoising techniques estimate pixel motions between adjacent frames via, for example, non-local matching~\cite{maggioni2012video,arias2019kalmancode,davy2019non,vaksman2021patch}, optical flow~\cite{xue2019video,yu2020jointcode,liang2022recurrent,buades2019enhancement}, kernel-prediction networks~\cite{mildenhall2018burst,xu2019learning,xu2020learning} and deformable convolutions~\cite{wang2019edvr}. 
Since accurate per-pixel motion estimation is challenging, methods with implicit motion modeling are also proposed~\cite{claus2019videnn,tassano2019dvdnet,bhat2021deep,mehta2021evrnet,tassano2020fastdvdnet,xiang2022remonet,maggioni2021efficient}.
These approaches either take simultaneously multiple frames as input, which are jointly processed by the model~\cite{claus2019videnn,tassano2019dvdnet,chen2021multiframe,sun2021deep,tassano2020fastdvdnet,yue2020supervised}, or use information from previous frames as the additional input~\cite{godard2018deep,ehmann2018real,maggioni2021efficient}.
Results of supervised methods on real noisy videos largely depend on the training data. 
Self-supervised video denoising~\cite{ehret2019modelcode,dewil2021self,sheth2021unsupervised,li2021learning} was explored. The result still falls behind that of supervised ones.

\section{Proposed PVDD Dataset}
\label{sec:pvdd}
Our PVDD, which is the main contribution of this work, contains 200 videos in both sRGB and RAW formats (captured with a NIKON Z7 \uppercase\expandafter{\romannumeral2} mirrorless camera), and each video clip spans 8s to 15s.

\vspace{-0.15in}
\subsection{Data Collection Setup}
For the RAW video collection, we use a NIKON Z7 \uppercase\expandafter{\romannumeral2} mirrorless camera and an Atomos Ninja \uppercase\expandafter{\romannumeral5} recording monitor.
The monitor is attached to the camera by an HDMI interface and is utilized to transport, encode and store the continuous video stream into an Apple ProRes RAW video (\textit{Apple ProRes RAW is a well-recognized video codec standard directly applied to the camera sensor's data}).
All videos are captured in resolution 1,920$\times$1,080@25p and 1,920$\times$1,080@60p format. The captured RAW videos can be further converted into sRGB videos by a calibrated camera ISP.

We capture the video by hand-held shooting to introduce additional camera motion with more complexity than conventional rig-based capture. We control the camera motion slowly to avoid visible motion jittering or motion blur. Camera setting, such as aperture, ISO, and focal length, is adjusted to minimize the noise level of the captured videos. For example, the aperture size is usually set between 4 and 6 to increase the amount of incident light received by the sensor, and the ISO is set below 400 to eliminate random noise. The focal length is adjusted by the camera's built-in auto-focus module to avoid out-of-focus blur.
Furthermore, we also compute the SNR for the videos in our PVDD dataset, obtaining the signal by denoising them by SOTA video denoising software, NeatVideo~\cite{neatvideo}. With the average SNR value as 48.46dB, it demonstrates that our captured videos can be safely considered as the ground truth.

\begin{figure*}[t]
    \centering
    \includegraphics[width=1.0\linewidth]{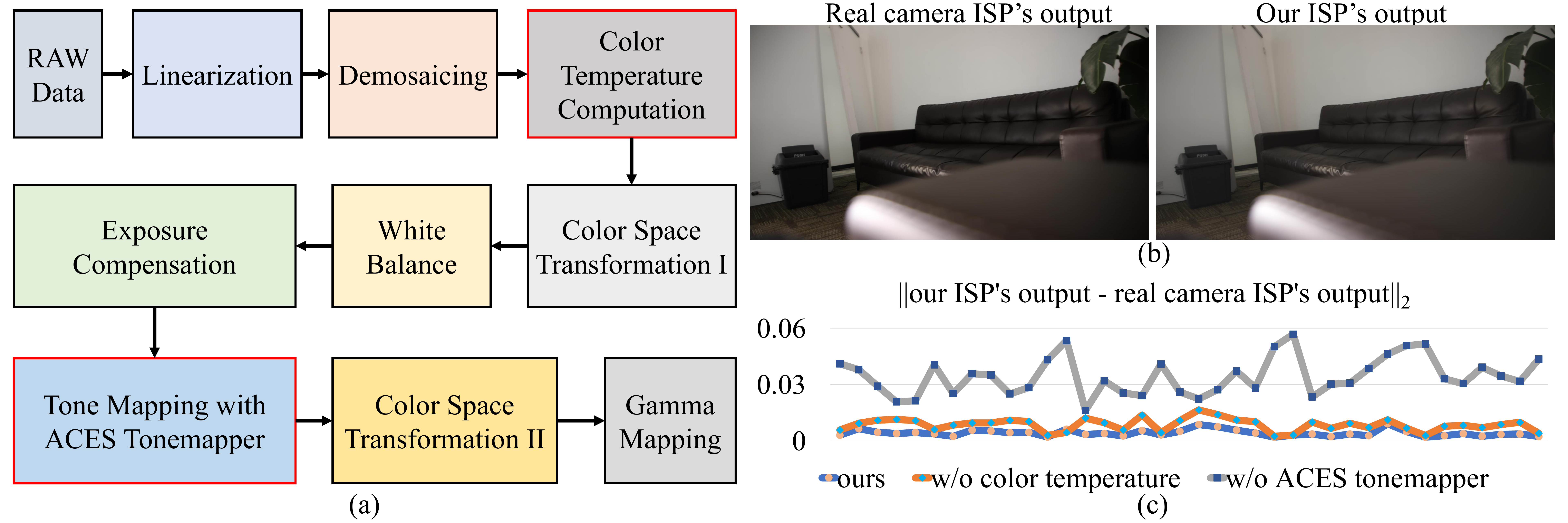}
    {
    \caption{
    (a) Modified ISP pipeline to produce sRGB videos. (b) Comparison between sRGB images produced by our modified synthetic ISP and the real camera ISP.
    More visual cases can be seen in Fig.~\ref{fig:sRGB222}.
        (c) Average normalized per-pixel $L_2$ distance between our ISP output and that of real camera ISP across 38 different scenes. Removing either the color temperature computation module or ACES tonemapper increases this distance.
    }
    \label{fig:sRGB}}
\end{figure*}

\begin{figure*}[t]
    \centering
    \includegraphics[width=1.0\linewidth]{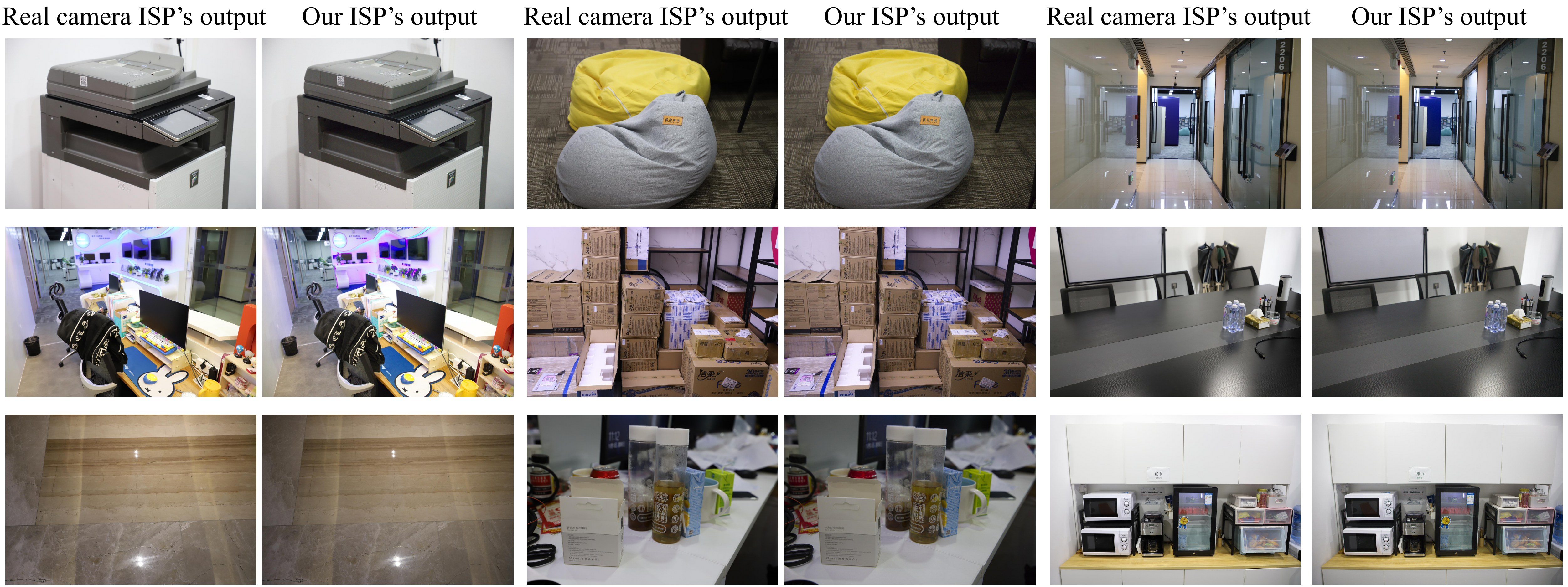}
    {
    \caption{The visualization to illustrate that our ISP's outputs are highly similar to the real camera ISP's output.}
    \label{fig:sRGB222}}
\end{figure*}

\subsection{Realistic Noise Synthesis}
\label{sec:synthesis}

\subsubsection{RAW Noise Synthesis} 
During collection, we capture one clean video for every scene, as shown in Fig.~\ref{fig:framework}.
We obtain a set of readout noise parameters $\sigma_r$ and shot noise parameters $\sigma_s$ at different ISO settings, which are pre-calibrated and recorded in the camera by the manufacturer.

To produce noisy RAW videos, we first randomly select ($\sigma_r^K$, $\sigma_s^K$) corresponding to a particular ISO $K$ and then add Gauss-Poisson noise accordingly to the clean RAW videos.
Suppose the clean RAW video is denoted as $\vec{Y}$. The process to obtain noisy observation $\vec{X}$ can be expressed by
\begin{equation}
\vec{X}_c \sim \mathcal{P}((\sigma_{s,c}^K)^2\vec{Y}_c)+\mathcal{N}(\vec{0}, (\sigma_{r,c}^K)^2),  c \in \{ \text{R, G, B} \},
\end{equation}
where $\mathcal{P}$ represents the Poisson distribution with mean $\vec{Y}_{c}$ and variance $(\sigma_{s,c}^K)^2 \vec{Y}_c$, and $\mathcal{N}$ denotes the Gaussian distribution with zero mean and variance $(\sigma_{r,c}^K)^2$.

The noise pattern in $\vec{X}$ is found close to real noises since 1) the noise distribution in the RAW domain can be well approximated by the Gauss-Poisson distribution~\cite{zhou2020awgn,yue2020supervised,abdelhamed2018highsidd}; 2) noise distribution model parameters are obtained from the actual calibration of the camera. 

\subsubsection{sRGB Video Pairs} 
To obtain noisy-clean video pairs in the sRGB domain, we adopt an integrated camera ISP pipeline as shown in Fig.~\ref{fig:sRGB}(a). Camera parameters in the ISP for each clean RAW video are provided by one additional noisy RAW image as shown in Fig.~\ref{fig:framework}.
Our ISP is built upon the ISP utilized in~\cite{abdelhamed2018highsidd} and our improvement is the following.

\subsection{Details of Our ISP}
In order to better approximate a real camera ISP, we 
1) compute the color temperature and serve it as a hyper-parameter to assist the color space transformation module with an accurate color transformation matrix; 2) choose an ACES tonemapper~\cite{reinhard2002photographic} as our tone mapping function and apply it to ProPhoto RGB color space, which has a wide color gamut~\cite{spaulding2000reference}. 
The example in Fig.~\ref{fig:sRGB}(b)\&(c) demonstrates the effectiveness of our improvement. 
    
In particular, the white balance module estimates an illumination value in terms of a color temperature, and the color space transformation I module (which maps the sensor-specific RAW RGB colors to a canonical space such as CIE XYZ) uses this estimated color temperature value to interpolate between two factory-calibrated color-correction matrices~\cite{rowlands2020color} to obtain an accurate matrix.

Moreover, as mentioned in~\cite{robertson1968computation}, the color transformation I module converts the cameraRGB space into CIE XYZ space, and then into ProPhoto RGB space. The color transformation II module converts ProPhoto RGB space into sRGB space.

\begin{figure*}[t]
    \centering
    \includegraphics[width=0.9\linewidth]{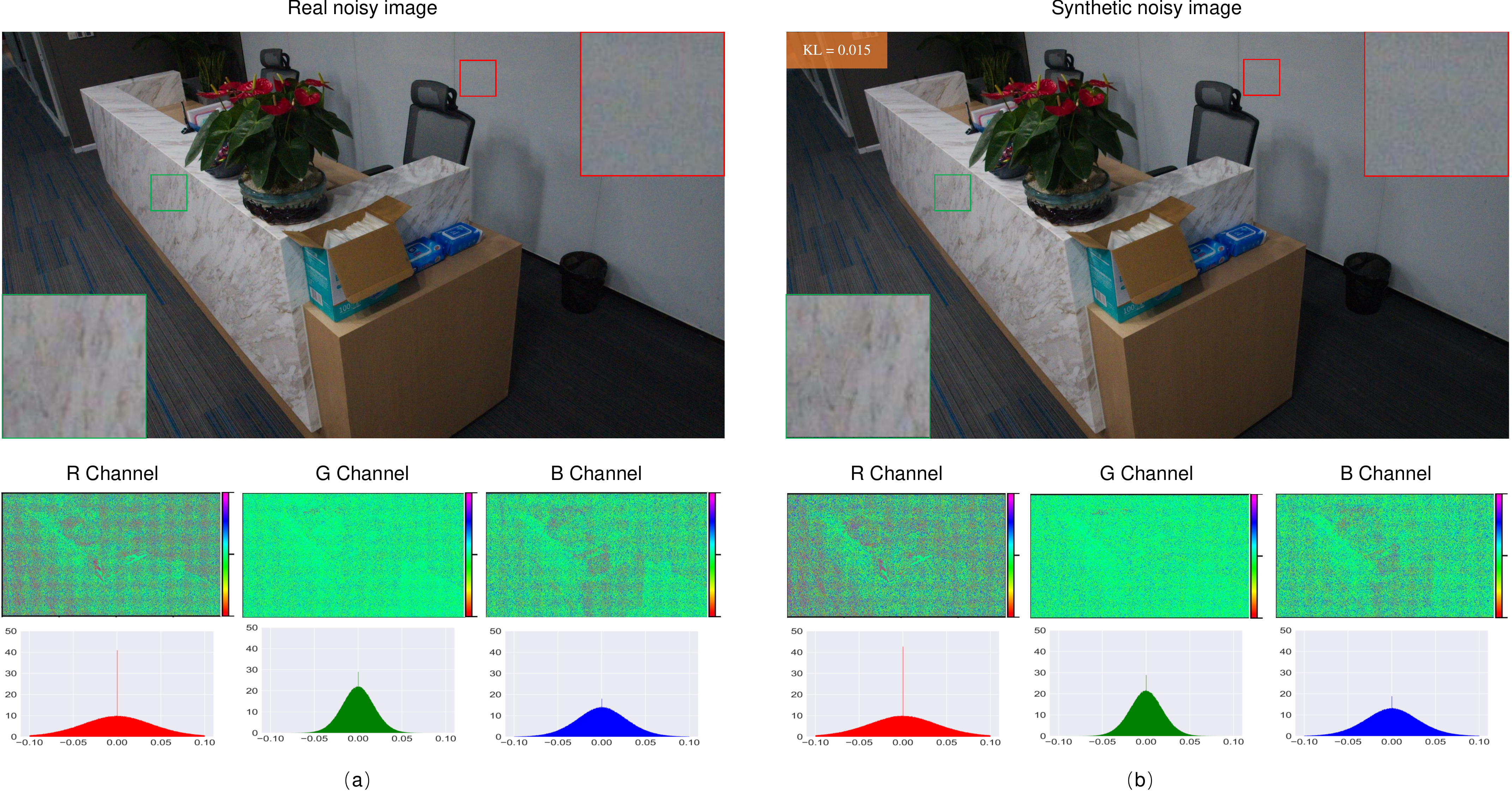}
    \caption{
        Illustration of similarity between the real and synthetic noise distributions in sRGB domain. 
    }
    \label{fig:noise_visualize}
\end{figure*}

\begin{figure*}[t]
    \centering
    \includegraphics[width=0.9\linewidth]{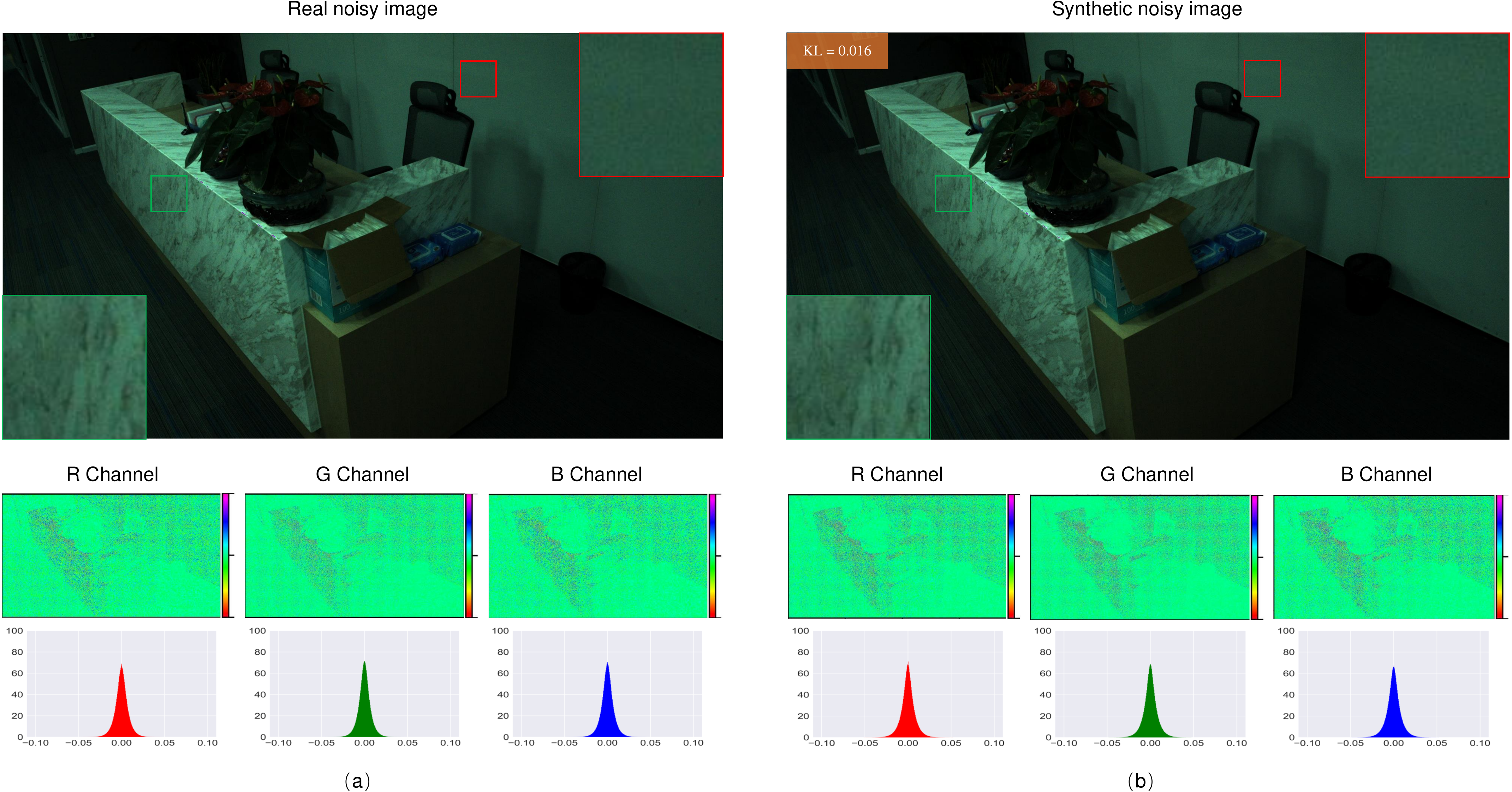}
    \caption{
        Illustration of similarity between the real and synthetic noise distributions in RAW domain. 
    }
    \vspace{-0.1in}
    \label{fig:noise_visualize_raw}
\end{figure*}

\begin{figure*}[t]
    \centering
    \includegraphics[width=1.0\linewidth]{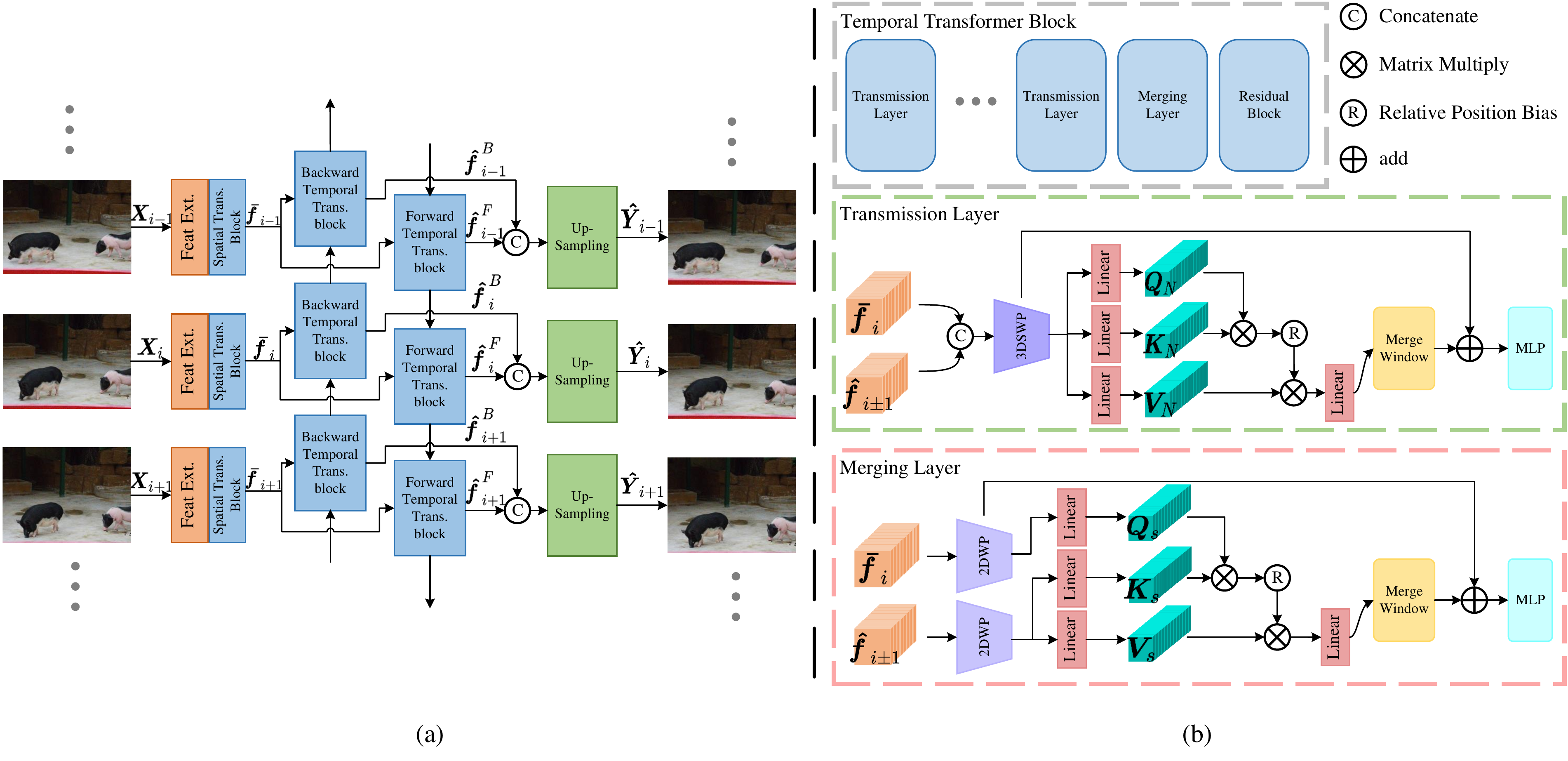}
    \caption{
        (a) The overview of the proposed RVDT, which consists of both spatial and bi-directional transformer blocks. (b) The details of the proposed bi-directional transformer blocks, which consist of both ``Transmission Layer" and ``Merging Layer".
    }
    \label{fig:pvdd-arch}
\end{figure*}

\subsection{Evaluation Dataset}
For the quantitative evaluation of our noise synthesis and the superiority over existing video denoising datasets, we also collect an evaluation dataset, a.k.a. \textit{Static15}, consisting of 15 different static scenes.
This dataset is produced by capturing static noisy videos by the tripod-mounted NIKON Z7 II camera, whereas the corresponding clean video is derived by averaging neighboring noisy video frames~\cite{nam2016holistic,yue2019high,brooks2019unprocessing,abdelhamed2018highsidd}.
Especially, we average 50 consecutive noisy frames to obtain a clean frame whose SNR value is 50.87dB proving clean frames' quality.

Note that we already generate noise on clean frames, and thus can compare it with the captured noise. 
As shown in Fig.~\ref{fig:noise_visualize}, the noise distribution of the synthetic noisy frame is surprisingly close to actual captured ones in the sRGB domain. 
We compute the KL divergence between synthesized and captured real noise in the sRGB domain across all video clips on Static15 (noise value range [0,1]). The average normalized per-pixel KL divergence is 0.034 over sRGB and 0.041 over RAW, indicating that the synthesized noise is realistic under the same camera setting (a KL divergence value below 0.05 is acceptable~\cite{abdelhamed2019noise}).
The noise distribution of the synthetic noisy frame is also very close to the real captured frame in the RAW domain, as displayed in Fig.~\ref{fig:noise_visualize_raw}.

\subsection{Editable Property of PVDD} 
Since PVDD provides videos in raw formats, it indicates that the whole simulation pipeline is highly editable. For example, existing noise models~\cite{zhang2021rethinking,chang2020learning,zhu2016noise,wang2019enhancing,nam2016holistic,wei2020physics} can also be utilized to create more sophisticated noise distributions in the RAW domain.
In addition, the parameters of the utilized ISP modules can be adjusted to make sRGB training data suitable for different target camera devices for future research.

\section{Recurrent Video Denoising Transformer}
\label{model}

Besides the newly built video denoising dataset, we further propose a novel video denoising framework with the spatial transformer and bi-directional temporal recurrent video transformer structure.
Our framework is called Recurrent Video Denoising Transformer (RVDT).
In this section, we first provide an overview of RVDT in Section~\ref{sec-arch}, and then we describe our newly designed attention blocks and feed-forward modules of RVDT in Section~\ref{sec-block} and Section~\ref{sec-mlp}, respectively.

\subsection{Model Architecture}
\label{sec-arch}

The significant components in video denoising are the propagation mechanisms in the spatial as well as temporal dimensions. 
To achieve the propagation, existing approaches can be divided into two categories, including the window-based and recurrent-based video denoising methods.
Window-based methods, e.g., FastDVDnet~\cite{tassano2020fastdvdnet} and RViDeNet~\cite{yue2020supervised}, typically take a video clip as the input and output the centered denoised frame. Although several models have been deliberately designed, recent works~\cite{chan2021basicvsr} have shown that they are computationally inefficient when the window size is large. Due to the limited input frame number, these window-based methods can only handle the propagation of short-term information among video frames.

On the other hand, recurrent-based methods are generally designed with high efficiency. They can capture long-term temporal information for denoising, formulated by propagating latent features sequentially for the input video clip.
Thus, one significant component in recurrent-based methods is the strategy for merging the adjacent latent features, since the inaccurate merging will lead to the accumulation of temporal errors.
One common strategy to merge is employing the computed optical flow~\cite{chan2021basicvsr} and deformable field~\cite{chan2022basicvsr++} between adjacent frames, which can propagate information frame-by-frame.
However, estimating precise motions from the noisy frames is difficult, suffering from the distortions and artifacts in videos. The imprecise estimation will lead to unsatisfactory results. 
Moreover, there are also some approaches that utilize the attention map between neighboring frames to build the inter-frame merging relation~\cite{maggioni2021efficient}.
Such attention mechanisms currently are mainly efficient for dealing with static scenes, while their effects on dynamic scenarios are limited.

Inspired by the tremendous success of the vision transformer's attention component in building non-local spatial correlations~\cite{wang2022deformable, liu2022video, geng2022rstt}, we in this paper propose to employ the transformer for formulating inter-frame temporal correlations in the recurrent-based video denoising framework. As shown in Fig.~\ref{fig:pvdd-arch}(a), without loss of generality, suppose we aim to complete the denoising process for the $i$-th frame $\vec{X}_i$.
The feature of $\vec{X}_i$, $\vec{\bar{f}}_i$, is extracted with the spatial CNN and transformer blocks, and it is merged with the features from temporally bi-directional features ($\vec{\hat{f}}_{i-1}^F$ or $\vec{\hat{f}}_{i+1}^B$) via our designed forward/backward temporal transformer blocks whose details are shown in Fig.~\ref{fig:pvdd-arch}(b).
With the non-local attention in the spatial dimension and the bi-directional attention in the temporal dimension, our recurrent-based video denoising framework is different from current video denoising approaches, can handle various motions efficiently, and achieve the SOTA denoising performance on all datasets (proved by the experiments in Sec.~\ref{sec:expersrgb}).

The denoising processing in RVDT consists of two stages.
In the first stage, as shown in Fig.~\ref{fig:pvdd-arch}(a), the feature of input frame $\vec{X}_i \in \mathbb{R}^{3\times H\times W}$ is extracted through a CNN-based encoder $\mathcal{C}_{S}$ as
\begin{equation}
    \vec{f}_i=\mathcal{C}_S(\vec{X}_i),
\end{equation}
where $\mathcal{C}_{S}$ consists of a shallow UNet and two convolutional layers with down-sampling operations, and $\vec{f}_i \in \mathbb{R}^{C_{in}\times \frac{H}{4}\times \frac{W}{4}}$.
The features extracted with a CNN-based encoder mainly involve local properties, while the denoising task also requires non-local operations for better performance.
Thus, we apply spatial 2D transformer blocks for $\vec{f}_i$, utilizing the corresponding attention computation module to find suitable non-local areas for the denoising of all image regions. The procedure can be formulated as
\begin{equation}
    \vec{\bar{f}}_i=\mathcal{T}_S(\vec{f}_i),
\end{equation}
where $\mathcal{T}_S$ denotes the concatenation of transformer blocks for spatial denoising.
The first stage acts as a spatial pre-denoiser, which benefits the temporal denoising in the second stage.

After acquiring the spatially denoised result $\vec{\bar{f}}_i$ from the first stage, it will be merged with bi-directional features ($\vec{\hat{f}}_{i-1}^F$ or $\vec{\hat{f}}_{i+1}^B$) which are formulated sequentially with the forward and backward temporal transformer blocks, as
\begin{equation}
\begin{array}{lcl}
    \vec{\hat{f}}_i^F  =  \mathcal{T}_T^F(\vec{\bar{f}}_i, \vec{\hat{f}}_{i-1}^F), \\
    \vec{\hat{f}}_i^B  =  \mathcal{T}_T^B(\vec{\bar{f}}_i, \vec{\hat{f}}_{i+1}^B),
\end{array}
\end{equation}
where $\vec{\hat{f}}_i^F$ and $\vec{\hat{f}}_i^B$ represent the forward and backward processed features, respectively.
$\mathcal{T}_T^F$ and $\mathcal{T}_T^B$ denote the forward and backward temporal transformer blocks.
In this way, the inter-frame correlations are built between $\vec{\bar{f}}_i$ and its adjacent features $\vec{\hat{f}}_{i\pm 1}$.

Finally, a decoder with up-sampling layers are employed to obtain the final denoised outputs, as
\begin{equation}
    \vec{\hat{Y}}_i = \mathcal{U}(\vec{\hat{f}}_i^F, \vec{\hat{f}}_i^B),
\end{equation}
where $\mathcal{U}$ is the decoder whose up-sampling layers are implemented with convolutions and pixel-shuffle layers~\cite{shi2016real}, and $\vec{\hat{Y}}_i$ is the denoised version of $\vec{X}_i$.

\subsection{Temporal Transformer Block}
\label{sec-block}

The details of the temporal transformer blocks will be described in this section. 
There are two kinds of strategies to implement the inter-frame attention via the transformer.
The first category employs the 3D transformer structure by conducting attention computations on 3D tokens to propagate information in both spatial and temporal dimensions implicitly.
Such transformer layers are called ``transmission layers".
The second approach views the bi-directional features as the references, and utilizes the 2D transformer to explicitly construct the relation between the current frame's feature and the references.
Such layers are called ``merging layers" that can merge neighboring frames' features.
The temporal transformer blocks in RVDT consist of these two kinds of layers.
In this way, the mutual feature propagation among neighboring frames is completed before the merge operation, and we find such a strategy can achieve better denoising performance.

Without loss of generality, we describe the details of the forward temporal transformer blocks, and the backward temporal transformer blocks are the same by changing $F$ to $B$, $i-1$ to $i+1$.
Suppose there are $N$ layers in $\mathcal{T}_T^F$, i.e., $\mathcal{T}_T^F=\{\mathcal{T}_{T,1}^F,...,\mathcal{T}_{T,N}^F \}$. Among them, $N$-1 layers are transmission layers, and the last one is the merging layer.
Thus, for the input of $\vec{\bar{f}}_i$ and $\vec{\hat{f}}_{i-1}^F$, there are $N$-1 intermediate outputs, as $\{\vec{\bar{f}}_{i,1},..., \vec{\bar{f}}_{i,N-1}\}$ and $\{  \vec{\hat{f}}_{i-1,1}^F,..., \vec{\hat{f}}_{i-1,N-1}^F \}$, respectively.
Thus, the procedure of the transmission layers is
\begin{equation}
\begin{array}{lcl}
\begin{cases}
    [\vec{\bar{f}}_{i,s}, \vec{\hat{f}}_{i-1,s}^F]  =  \mathcal{T}_{T,s}^F(\vec{\bar{f}}_i, \vec{\hat{f}}_{i-1}^F),  \quad \quad \quad \, \, \, s=1 \\
    [\vec{\bar{f}}_{i,s}, \vec{\hat{f}}_{i-1,s}^F]  =  \mathcal{T}_{T,s}^F(\vec{\bar{f}}_{i,s-1}, \vec{\hat{f}}_{i-1,s-1}^F), \, s\in[2, N-1] \\
\end{cases}.
\end{array}
\end{equation}
The features obtained from the transmission layers are processed with the merging layer to acquire the final output of the temporal transformer blocks, as
\begin{equation}
\vec{\hat{f}}_{i}^F  =  \mathcal{T}_{T,N}(\vec{\bar{f}}_{i,N-1}, \vec{\hat{f}}_{i-1,N-1}^F).
\end{equation}
The implementation details of the transmission layers and the merging layer are described in the following.

\noindent {\bfseries Transmission Layer.}
The overall architecture of the transmission layer is shown in the middle part of Fig.~\ref{fig:pvdd-arch}(b).
For the input of $[\vec{\bar{f}}_{i,s}, \vec{\hat{f}}_{i-1,s}^F], s\in [1,N-1]$, they are concatenated and processed with 3D window partition layers ($\text{3DWP}$) to serve as the input of 3D transformer layers.

For the window-based attention computation in the $s$-th transmission layer, the query $\vec{Q}_{s}$, key $\vec{K}_{s}$, value $\vec{V}_{s}$ vectors are obtained via linear projections, as
\begin{equation}
\begin{aligned}
    \vec{h}_s &= \text{3DWP}([\vec{\bar{f}}_{i,s}, \vec{\hat{f}}_{i-1,s}^F]),\\
    \vec{Q}_{s} &= \vec{W}^s_q \cdot \text{LN}(\vec{h}_s),\\
    \vec{K}_{s} &= \vec{W}^s_k \cdot \text{LN}(\vec{h}_s), \\
    \vec{V}_{s} &= \vec{W}^s_v \cdot \text{LN}(\vec{h}_s),
\end{aligned}
\end{equation}
where $[\cdot,\cdot]$ denotes the concatenation operation, $\text{LN}$ represents the layer normalization~\cite{ba2016layer}, $\vec{W}^s_q$, $\vec{W}^s_k$, and $\vec{W}^s_v$ are matrices of learnable parameters for linear projections.
The attention output can be written as
\begin{equation}
\label{eq-attn}
\text{Attention}(\vec{Q}_s, \vec{K}_s, \vec{V}_s) = \text{SoftMax}(\vec{Q}_s\vec{K}_s^T / \sqrt{d_s} + \vec{B}_s)\vec{V}_s,
\end{equation}
where $d_s$ is for normalization in the $s$-th layer, and $\vec{B}_s$ is the relative position bias.
The feature tokens are updated as
\begin{equation}
\vec{h}'_s=\vec{h}_s+\text{Attention}(\vec{Q}_s, \vec{K}_s, \vec{V}_s).
\end{equation}

Moreover, in the feed-forward module of the transformer, we adopt a channel-spatial attention MLP (CSA-MLP), which is denoted as $\mathcal{M}$ and will be described in the next section. And the feature tokens after the 3D window reverse layers (3DWR) can formulate the output of the $i$-th transmission layer. The overall process can be formulated as
\begin{equation}
\label{eq:mlp1}
[\vec{\bar{f}}_{i, s+1}, \vec{\hat{f}}_{i-1, s+1}^F] = \text{3DWR}(\mathcal{M}(\text{LN}(\vec{h}'_s))+\vec{h}'_s).
\end{equation}

\noindent {\bfseries Merging Layer.} 
Within the merging layer, the input feature $\vec{\bar{f}}_{i,N-1}$ and $\vec{\hat{f}}_{i-1,N-1}^F$ are processed with the 2D window partition layer (2DWP) separately to obtain the corresponding feature tokens.
The target of the merging layer is to merge the bi-directional features, i.e., $\vec{\hat{f}}_{i-1,N-1}^F$, to the feature of the current frame, i.e., $\vec{\bar{f}}_{i,N-1}$.
This is achieved by the inter-frame relationship computed in window-based attention module in the merging layer, which can be written as
\begin{equation}
\begin{aligned}
\vec{m}_N &= \text{2DWP}(\vec{\bar{f}}_{i,N-1}),\\
\vec{n}_N &= \text{2DWP}(\vec{\hat{f}}_{i-1,N-1}^F),\\
    \vec{Q}_{N} &= \vec{W}^N_q \cdot \text{LN}(\vec{m}_N),\\
    \vec{K}_{N} &= \vec{W}^N_k \cdot \text{LN}(\vec{n}_N), \\
    \vec{V}_{N} &= \vec{W}^N_v \cdot \text{LN}(\vec{n}_N),
\end{aligned}
\end{equation}
where $\vec{W}^N_q$, $\vec{W}^N_k$, and $\vec{W}^N_v$ represent the projection matrices in the merging layer's attention module.
In this way, each feature token can be merged with the information from the temporally bi-directional features, as
\begin{equation}
\vec{m}'_N=\vec{m}_N+\text{Attention}(\vec{Q}_{N}, \vec{K}_{N}, \vec{V}_{N}).
\end{equation}
The feature acquired from the attention module is processed with the CSA-MLP and 2D window reverse layer (2DWR) to obtain the final output, as
\begin{equation}
\label{eq:mlp2}
\vec{\hat{f}}_i^F=\text{2DWR}(\mathcal{M}(\text{LN}(\vec{m}'_N))+\vec{m}'_N).
\end{equation}

\subsection{Channel-Spatial Attention MLP Block}
\label{sec-mlp}

\begin{figure}[t]
    \centering
    \includegraphics[width=1.0\linewidth]{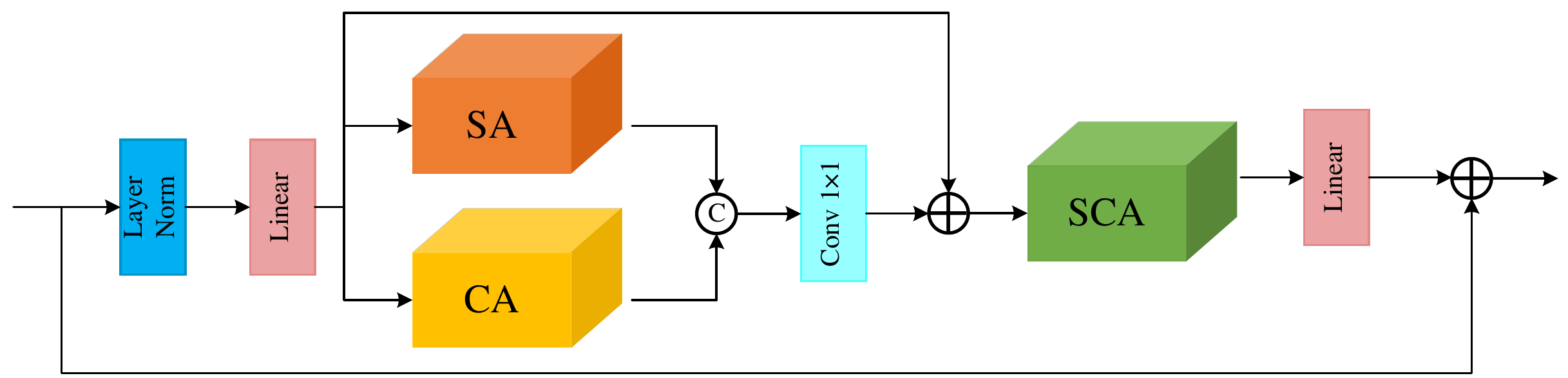}
    \caption{
    The architecture of the proposed Channel-Spatial Attention MLP Blocks in transformers' feed forward stages.
    }
    \label{fig:mlp-arch}
\end{figure}

Generally, the feed-forward module in the transformer is formulated as an MLP~\cite{liu2021swin}, enhancing the representation capacity of feature tokens.
Recently, more and more works have noticed the drawbacks of modeling local relations via an MLP in the feed-forward module, and extra modules are introduced to enhance the local property, e.g., the CNN blocks~\cite{liang2021swinir}.
In this work, we further propose a channel-spatial attention MLP block to formulate the feed-forward module.
As shown in Fig.~\ref{fig:mlp-arch}, the key points of this design are twofold: (a) we keep the original MLP to capture the long-range dependencies, (b) we incorporate the attention-based convolutional layers to capture short-range dependencies. Different from the existing locally enhanced MLP module~\cite{liang2021swinir}, channel attention (CA), spatial attention (SA), and jointly spatial-channel attention (SCA) are inserted in the convolutional layers to better model the local dependencies, benefiting the denoising effects.

Given the feature $\vec{z}$ from the attention module, i.e., $\vec{h}'_s$ in Eq.~\ref{eq:mlp1} and $\vec{m}'_N$ in Eq.~\ref{eq:mlp2}.
The overall procedure in the CSA-MLP can be formulated as
\begin{equation}
\begin{array}{lcl}
    \vec{\bar{z}}  =  \mathcal{M}_1(\text{LN}(\vec{z})), \\
    \vec{\bar{z}} = \text{Reshape}(\vec{\bar{z}}),\\
    \vec{\hat{z}}  =  \text{SCA}(\text{Conv}([\text{SA}(\vec{\bar{z}}), \text{CA}(\vec{\bar{z}})]) + \vec{\bar{z}}) + \vec{\bar{z}}, \\ 
    \vec{\hat{z}} = \text{Flatten}(\vec{\hat{z}}),\\
    \vec{\hat{z}} = \mathcal{M}_2(\vec{\hat{z}}) + \vec{z},
\end{array}
\end{equation}
where $\mathcal{M}_1(\cdot)$ and $\mathcal{M}_2(\cdot)$ are two MLPs in Fig.~\ref{fig:mlp-arch} to construct long-range relations, $\text{Conv}(\cdot)$ is a convolutional layer, $\text{Reshape}$ is the operation to turn feature tokens with the shape as $(B\times T) \times C \times (h \times w)$ into the 2D maps with the shape as $(B\times T) \times C \times h \times w$ ($B$ is the batch size, $T \in \{1, 2\}$ is the temporal size, $C$ is the channel number, $h$ and $w$ are the height and the width, respectively), and $\text{Flatten}$ is the reverse process of $\text{Reshape}$.

SA exploits the spatial relationships of features, and CA is designed to build the inter-channel dependencies of convolutional features.
For the input with shape as $(B\times T) \times C \times h \times w$, the attention map from SA is $(B\times T) \times 1 \times h \times w$, and from CA is $(B\times T) \times C \times 1 \times 1$.
The attention value is depicted with the sigmoid function.
Furthermore, to better extract the content information, we jointly optimize the attention mechanisms in the spatial-channel dimensions via SCA, which is not considered by previous works.
The attention map from SCA has the shape as $(B\times T) \times C \times h \times w$, which can refine the features processed with the locally enhanced convolutional layer.

\subsection{Implementation Details}
We implement our RVDT in PyTorch~\cite{paszke2019pytorch}, and train and test it on a PC with an NVIDIA 2080Ti GPU. 
The proposed RVDT is trained with the reconstruction loss between the output and the noise-free ground truth.
For both sRGB and RAW domains, training patches with a size of 256$\times$256 are extracted randomly as inputs, and we adopt common augmentations, e.g., rotation and flip.
Especially for the RAW domain, we extract patches after packing one-channel Bayer RAW images into four-channel GBRG images, to avoid sampling improper patterns~\cite{liu2019learning}. We use Adam optimizer~\cite{kingma2014adam} with momentum set to 0.9. The batch size and learning rate are set as 1 and $1\times10^{-4}$ for training, respectively. The RVDT model is trained until convergence.
\textit{We will release our dataset and code after publication to benefit the research community.}

\section{Experiments}
\label{sec:expersrgb}
In this section, we first describe the data, networks, and metric to evaluate the effect of PVDD and RVDT in Sec.~\ref{dataset-compare}, Sec.~\ref{sec:denosing_networks}, and Sec.~\ref{sec:exp_metrics}, respectively.
We train the same network structure on different datasets to illustrate the superiority of our PVDD dataset over existing video denoising benchmarks in Sec.~\ref{sec:eva_dataset}; we conduct experiments with different video denoising networks on the same dataset to demonstrate the effectiveness of our proposed RVDT in Sec.~\ref{sec:eva_network}.

\begin{figure*}[t]
    \centering
    \includegraphics[width=1.0\linewidth]{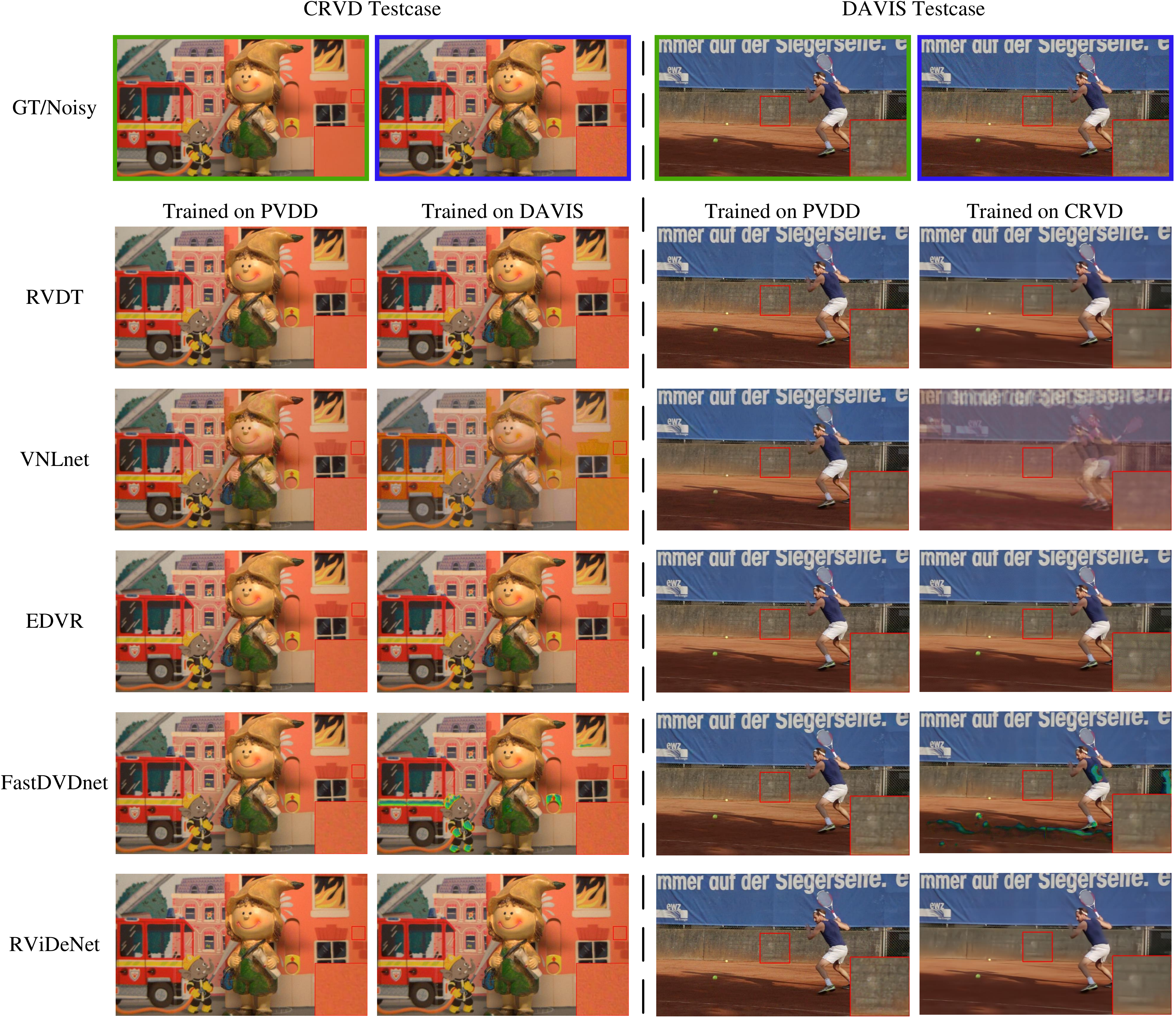}
    \caption{
Visual comparison with varying video denoising networks under the cross-dataset evaluation setting. “Noisy” is the noisy input, and ``GT" is the ground truth. The model trained with PVDD can be better generalized for denoising videos in the wild that have never appeared during the training, yielding superior visual results on both the smooth area and the region with complex texture.
    }
    \label{fig:dataet-cross}
\end{figure*}

\begin{figure}[t]
    \centering
    \includegraphics[width=1.0\linewidth]{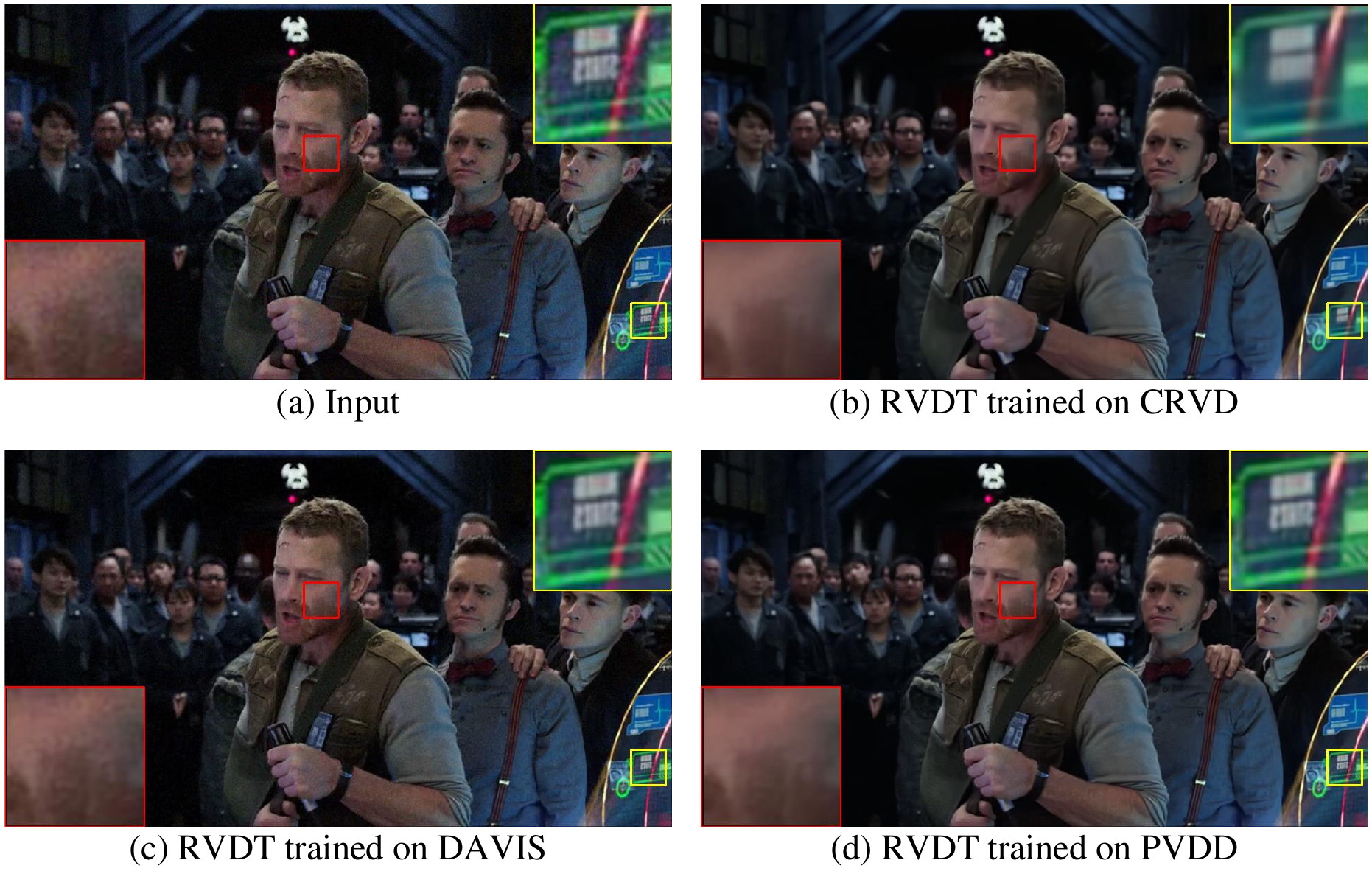}
    \caption{
    The denoising results towards real-world noisy videos for RVDT trained on different video denoising datasets and tested on General15. The network trained on PVDD gives the most pleasing results: noise-free smooth area (red rectangle); clean, realistic, and sharp denoised results for regions with complex textures (yellow rectangle).
    }
    \label{fig:dataset-poc-rvdt}
\end{figure}

\begin{figure}[t]
    \centering
    \includegraphics[width=1.0\linewidth]{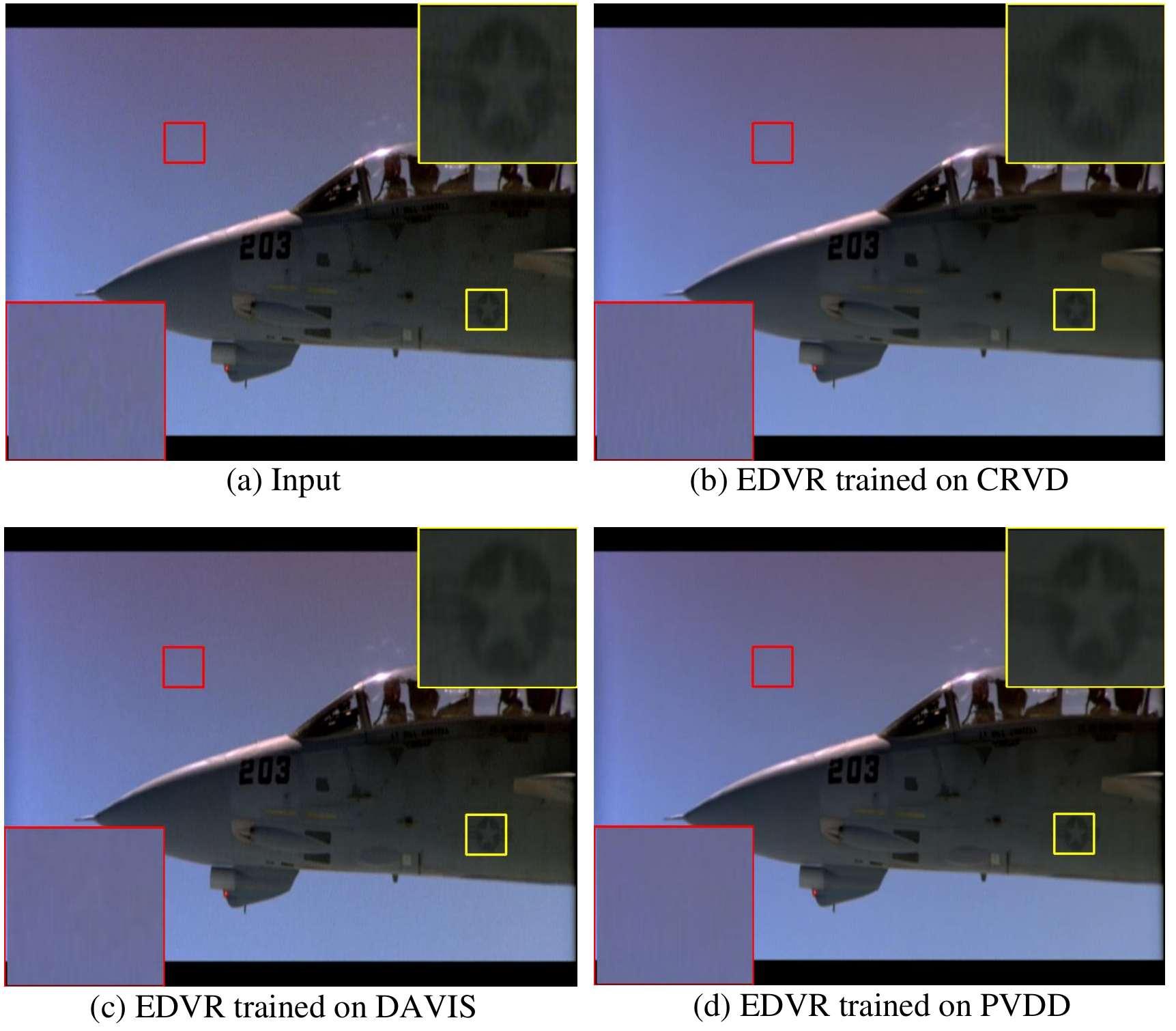}
    \caption{
    The denoising results towards real-world noisy videos for EDVR trained on different video denoising datasets and tested on General15. The network trained on PVDD gives the most pleasing results: noise-free smooth area (red rectangle); clean, realistic, and sharp denoised results for regions with complex textures (yellow rectangle).
    }
    \label{fig:dataset-poc-edvr}
\end{figure}

\subsection{Datasets}
\label{dataset-compare}
We first illustrate the training and testing datasets that will be utilized in the evaluation of PVDD and RVDT.

\subsubsection{Training data} We train representative state-of-the-art video denoising networks on both PVDD and popular benchmark datasets and compare the performance under the same network structure or dataset.
Previous datasets include CRVD~\cite{yue2020supervised} (including synthesized data part~\cite{milan2016mot16}) and DAVIS~\cite{perazzi2016benchmark}.
Note that low-light video datasets, e.g., SMID~\cite{chen2019seeing} and SDSD~\cite{wang2021seeing,xu2022snr} datasets, are not considered since their noisy and clean frames have different illumination values, making training difficult.

We divide all the videos in PVDD into the training collection (90\%) and the testing collection (10\%). As it is not necessary to employ all the frames (71,714 in total)
for training, for each video in the collection, we select 2 or 3 clips (each contains 25 continuous frames) randomly to form the actual training set. 

As for CRVD, RAW data is transferred to the sRGB domain using the default ISP implemented by SID~\cite{chen2018learning}. 
As for DAVIS, we first use the unprocessing technique~\cite{brooks2019unprocessing} to turn sRGB videos into the RAW format, and add Gauss-Poisson noise in the RAW domain to form pairs. The noisy RAW videos are finally converted back to sRGB. This is a common setting adopted by \cite{tassano2020fastdvdnet,lee2021restore}.

\subsubsection{Evaluation data} 
We consider the evaluation setting, where noisy sRGB videos are obtained from ISP without post-processing, which is the common setting for current works~\cite{yue2020supervised,tassano2020fastdvdnet}.

Specifically, the testing dataset of PVDD has two types of data.
The first is selected from the testing set of PVDD to contain only dynamic scenes. 
20 dynamic clean video clips are picked from the testing set of PVDD to form \textit{Dynamic20}.
For each clip, three noisy versions are generated with noise levels of ``Heavy", ``Medium", and ``Light", corresponding to ISO of 20,000, 8,000, and 2,500. 
The second type is \textit{Static15} (as explained in Sec.~\ref{sec:pvdd}), which consists of 15 static videos with real camera noise.
These noisy videos are captured with 1,920$\times$1,080 resolution at 60 FPS under various ISO settings ranging from 16,000 to 25,600. The corresponding clean version is directly obtained by averaging consecutive noisy frames. 

Moreover, we employ the testing dataset of CRVD~\cite{yue2020supervised} and DAVIS~\cite{perazzi2016benchmark} to evaluate the effectiveness of our proposed video denoising dataset and network.

\begin{table}[t]
    \centering
    \huge
\caption{Quantitative comparison for training on PVDD or CRVD/DAVIS dataset, while evaluating on DAVIS/CRVD dataset.}
\label{comparison-dataset}
\resizebox{1.0\linewidth}{!}{
    \begin{tabular}{l|p{4.5cm}<{\centering}|p{4.5cm}<{\centering}|p{4.5cm}<{\centering}|p{4.5cm}<{\centering}}
        \toprule[1pt]
        &\multicolumn{2}{c|}{eval. DAVIS}&\multicolumn{2}{c}{eval. CRVD}\\
        \cline{2-5}
        & train. PVDD & train. CRVD & train. PVDD & train. DAVIS \\
        \hline
        VNLnet & 32.83/0.912 & 18.19/0.580  & 29.61/0.855 & 26.20/0.850 \\
        FastDVDnet & 35.96/0.935 & 21.08/0.726  & 32.33/0.807 & 30.12/0.801  \\
        RViDeNet & 37.06/0.950 & 33.85/0.891    & 32.34/0.810 & 33.63/0.819  \\
        EDVR & 36.89/0.943 & 31.84/0.863        & 34.89/0.870 & 32.75/0.796  \\
        RVDT & 37.30/0.950 & 33.11/0.870       & 34.99/0.870 & 32.59/0.789  \\
        \bottomrule[1pt]
\end{tabular}}
\end{table}

\begin{figure*}[t]
    \centering
    \includegraphics[width=1.0\linewidth]{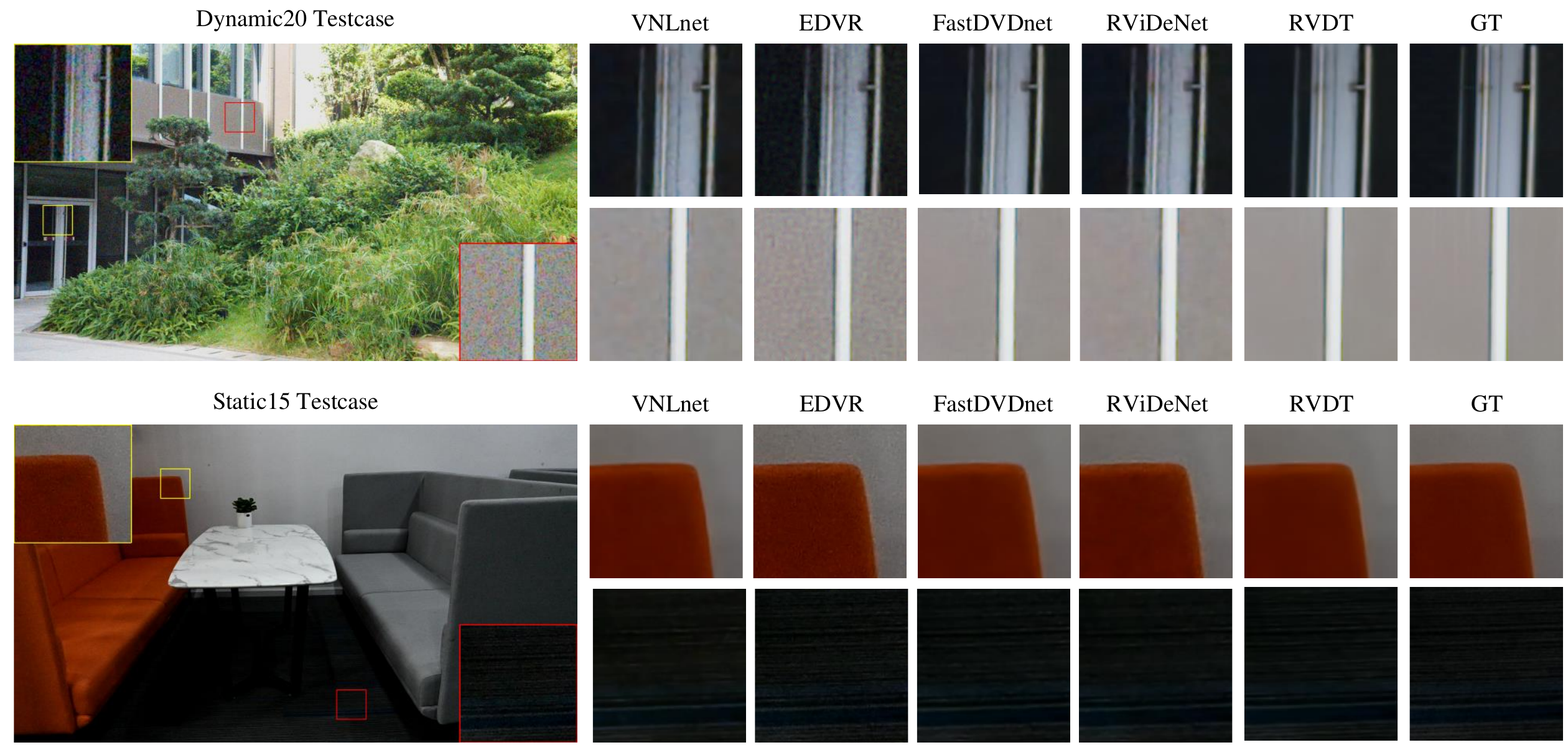}
    \caption{
The visual comparison among different networks trained on PVDD. These networks are trained in the sRGB domain with the blind version. Our designed RVDT has better denoising effects towards various regions.
    }
    \label{fig:network-srgb-blind}
\end{figure*}

\subsection{Denosing Networks}
\label{sec:denosing_networks}
We choose current SOTA video denoising frameworks for experiments, demonstrating the effects of our dataset and RVDT in Sec.~\ref{sec:eva_dataset} and Sec.~\ref{sec:eva_network}, respectively.
These denoising frameworks include the window-based and recurrent-based approaches, and they
are released with publicly available training and evaluation codes, 
including 
VNLnet~\cite{davy2019non}, FastDVDnet~\cite{tassano2020fastdvdnet},
EDVR~\cite{wang2019edvr}, RViDeNet~\cite{yue2020supervised}, and EMVD~\cite{maggioni2021efficient} (EMVD is utilized for the evaluation of RAW only that is the target of EMVD's original paper).

We follow the original setup proposed by the authors to train these networks, i.e., 
15 consecutive frames for VNLnet, 5 for FastDVDnet and EDVR, and 3 for RViDeNet, respectively. We use the official implementation for training and evaluation, and all training configurations remain as the default ones. 
We train both blind (without noise level prior) and non-blind (with noise level prior) versions for each dataset-network setup. 
For the models with noise levels as inputs, we set the noise level map as an additional input channel for each frame. Thus, input channel numbers are different for blind and non-blind models.

\subsection{Metrics}
\label{sec:exp_metrics}
For the evaluation on datasets with ground truth (e.g., Dynamic20 and Static15), we compute the PSNR and SSIM~\cite{wang2004image} between model output and ground truth.

\subsection{Evaluation for Denoising Dataset}
\label{sec:eva_dataset}
In this section, we compare PVDD with existing two video denosing datasets, including CRVD and DAVIS. We aim to prove the superiority of PVDD in training video denoising networks.

\subsubsection{Evaluation Strategy}
To demonstrate the superiority of our framework and avoid the bias of the evaluation datasets, we set a cross-dataset evaluation strategy. For example, to demonstrate the superiority of PVDD over CRVD, we train different networks on PVDD and CRVD, respectively, and evaluate the performance of trained networks on another dataset, i.e., DAVIS.
In the similar way, for the comparison between PVDD and DAVIS, we adopt CRVD as the evaluation dataset.

\subsubsection{Quantitative Results}
As shown in the left part of Table~\ref{comparison-dataset} (we still employ the practical setting where networks are evaluated with the blind version in the sRGB domain), the networks trained on PVDD can be better generalized to the denoising on the DAVIS dataset than the models trained on the CRVD. For example, compared with RVDT trained on the datasets of CRVD, the same model trained on our dataset achieves more than \textit{4 dB} gain on DAVIS (in terms of PSNR), which is significant. The advantage of SSIM is also prominent.
The advantage stems from the large scale of PVDD that contains natural motion and realistic noise.

Moreover, the generalization superiority is still apparent when the comparison is conducted between PVDD and DAVIS, employing the testing set of CRVD for evaluation. The results are displayed in the right part of Table~\ref{comparison-dataset}.
Models trained on PVDD can reach 32.90/0.842 (PSNR/SSIM) on average on the CRVD testing dataset, while the models trained on DAVIS can only achieve 31.06/0.811 (PSNR/SSIM) on average, verifying the usefulness of diverse motion, realistic noise, and various scenes in our PVDD dataset.

All these improvements are achieved without bias from the training data, guaranteeing the worth in measuring the denoising performance on real-world videos. Thus, in summary, the model trained on our PVDD dataset yields a better denoising generalization ability.

\subsubsection{Qualitative Results}
We provide the visual comparisons for the results in Table~\ref{comparison-dataset}.
In Fig.~\ref{fig:dataet-cross}, we show the visual comparison with the cross-dataset evaluation setting. The two left columns are the results from the networks tested on CRVD, and trained on PVDD and DAVIS, respectively. Obviously, the model trained with PVDD yields more excellent visual quality. For example, in the areas of red rectangles in Fig.~\ref{fig:dataet-cross}, the networks trained with PVDD generate the more pleasing result with most of the noise eliminated.
Similarly, the two right columns of Fig.~\ref{fig:dataet-cross} provide the visual comparison for networks tested on DAVIS, and trained with PVDD and CRVD, respectively. 
It can be seen that the models trained with our PVDD give more satisfactory results than the networks trained with CRVD. The superiority can be observed in both smooth areas (most noise removed) and regions with rich textures (details preserved). 
The excellent generalization ability of the model trained with PVDD comes from the large-scale data and the realistic noise distribution in PVDD.
Moreover, as shown in the right two columns of Fig.~\ref{fig:dataet-cross}, the VNLnet trained with CRVD could have motion blurry artifacts when tested on DAVIS, since the motion distribution in CRVD is much smaller than the DAVIS with natural moves. On the other hand, our PVDD provides various motion distributions to avoid such blurry artifacts.

\subsubsection{Real-world Denoising Evaluation}
In the real world, a noisy sRGB video from the ISP is very likely to be further processed with a post-processing pipeline in practice, i.e., noise in the sRGB domain is further distorted by various types of degradation, e.g., UV denoiser~\cite{tomasi1998bilateral,hasinoff2016burst}, blur~\cite{liu2020estimating}, compression~\cite{cheng2022optimizing}, resizing~\cite{nataraj2009adding}, etc.
To further evaluate the generalization of models trained on competing datasets for noisy videos undergoing complicated real-world degradations, we select 15 noisy videos from the Internet that undergo various real-world post-processing operations in different applications to form \textit{General15}.
Each video in General15 suffers from noise contamination from the blind post-processing pipeline.

In Figs.~\ref{fig:dataset-poc-rvdt} and~\ref{fig:dataset-poc-edvr}, we show a visual comparison of test videos in the General15 dataset with
practical noise. The model trained with PVDD yields more outstanding visual quality. For example, RVDT in Fig.~\ref{fig:dataset-poc-rvdt} trained with PVDD generates the most pleasing results in both smooth areas (most noise eliminated) and regions with rich textures (details preserved). Results from training with the CRVD dataset show
blurry artifacts, while those with the DAVIS dataset retain much noise.
The superiority of PVDD profits from its realistic noise, practical motion patterns, and diverse scenes.
This evaluation set, General15, will also be publically released to the research community as a real-world video denoising benchmark.

\begin{table*}[t]
    \centering
\large
\caption{Quantitative comparison in both the sRGB and RAW domain on PVDD dataset, -RAW means the RAW models. We report PSNR/SSIM}
\label{comparison-rvdt-pvdd}
\resizebox{1.0\linewidth}{!}{
    \begin{tabular}{l|p{3.8cm}<{\centering}|p{3.8cm}<{\centering}|p{3.8cm}<{\centering}|p{3.8cm}<{\centering}}
        \toprule[1pt]
        &\multicolumn{2}{c|}{Dynamic20}&\multicolumn{2}{c}{Static15}\\
        \cline{2-5}
&blind & non-blind & blind & non-blind \\
\hline
VNLnet&33.30/0.912&33.69/0.918&35.33/0.933&35.89/0.926\\
FastDVDnet&34.46/0.916&34.78/0.923& 35.25/0.908&36.09/0.929\\
RViDeNet&33.90/0.911&34.15/0.915&35.31/0.919&35.98/0.917\\
EDVR&31.34/0.789&31.78/0.810&33.98/0.866&34.46/0.877\\
RVDT&\textbf{35.71/0.941}&\textbf{35.85/0.942}&\textbf{35.76/0.924}&\textbf{36.61/0.941}\\
        \hline
        \hline
        VNLnet-RAW & 37.05/0.941 & 35.75/0.933 & 42.21/0.980 & 41.01/0.975 \\
        FastDVDnet-RAW & 43.95/0.966 & 42.17/0.965 & 46.42/0.983 & 47.94/0.989\\
        RViDeNet-RAW & 43.92/0.977 &43.76/0.975 & 47.14/0.985 & 47.95/0.984 \\
        EDVR-RAW & 36.60/0.873 & 36.66/0.875 & 42.57/0.954 & 42.64/0.955 \\
        EMVD-RAW & 39.15/0.934 & 38.44/0.928 & 46.16/0.983 & 44.82/0.978\\
        RVDT-RAW & \textbf{44.30/0.979} & \textbf{44.86/0.981} & \textbf{48.70/0.992} & \textbf{48.56/0.991} \\
        \bottomrule[1pt]
\end{tabular}}
\end{table*}

\begin{figure*}[t]
    \centering
    \includegraphics[width=1.0\linewidth]{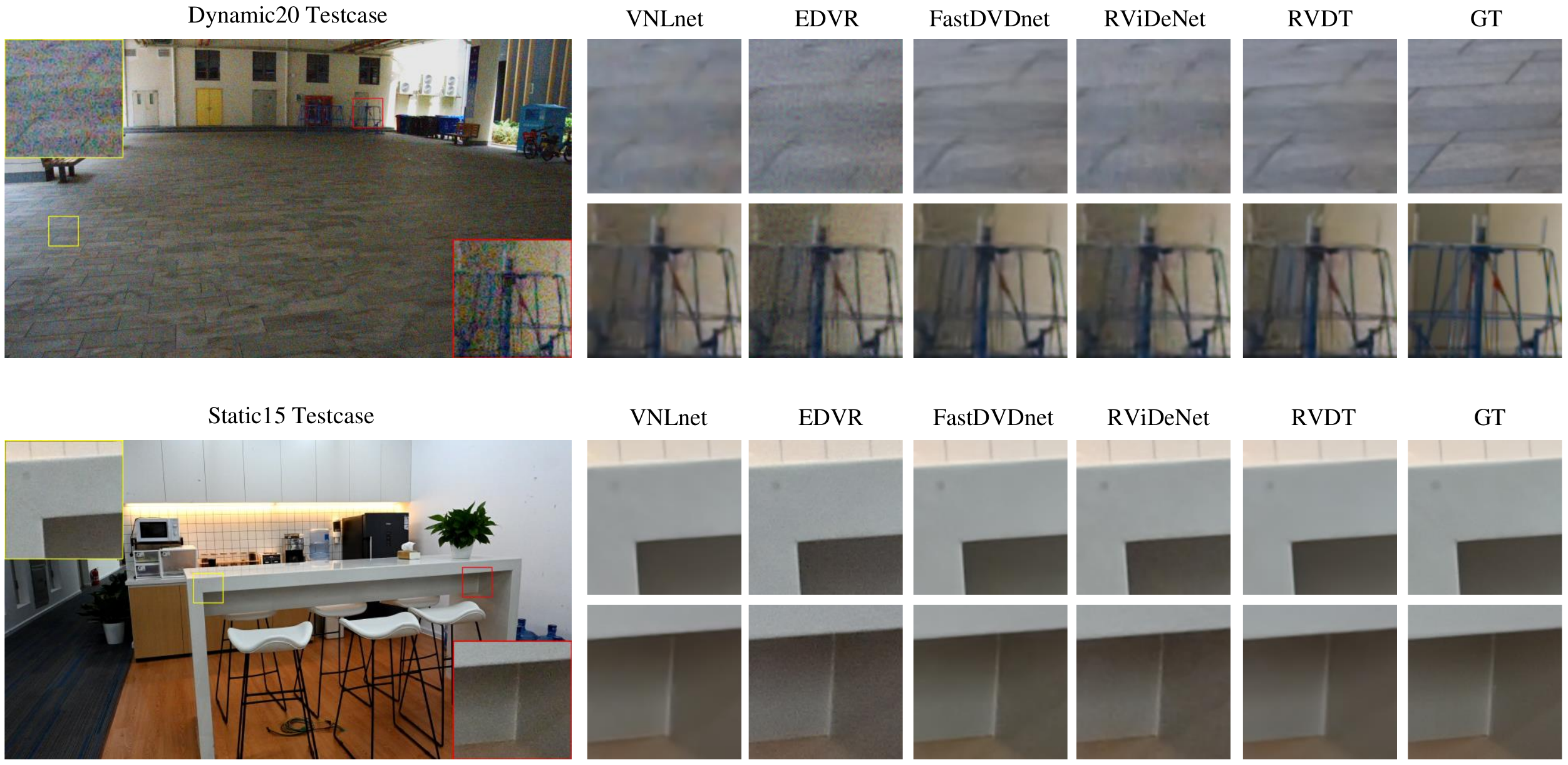}
    \caption{
    The visual comparison for different video denoising frameworks trained on PVDD. These networks are trained in the sRGB domain with the non-blind version. Our framework has the best denoising performance in terms of noise removal and detail preservation.
    }
    \label{fig:network-srgb-noblind}
\end{figure*}

\begin{figure*}[t]
    \centering
    \includegraphics[width=1.0\linewidth]{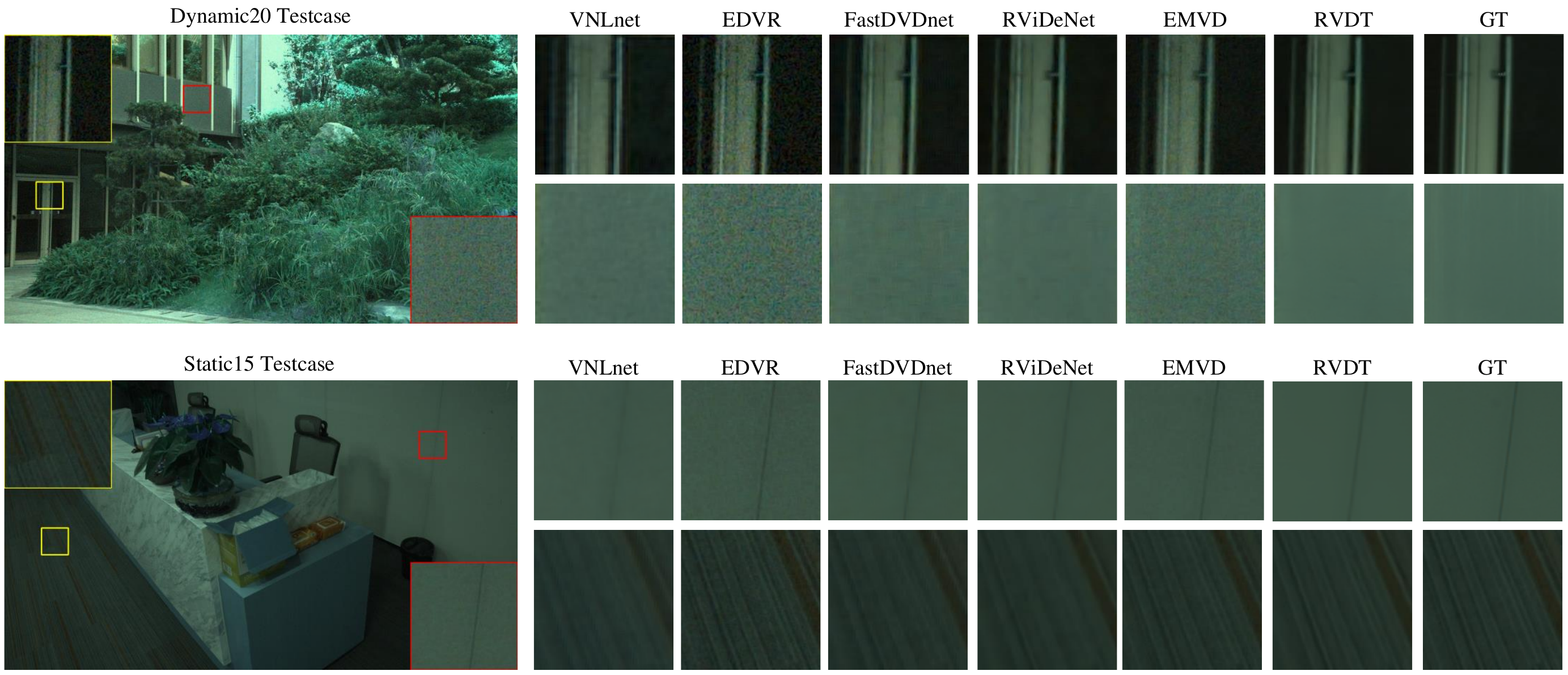}
    \caption{
    The qualitative comparison among different video denoising models trained on PVDD, which are trained in the RAW domain with the non-blind version. Our framework achieves the best denoising effects.
    }
    \label{fig:network-raw-blind}
\end{figure*}

\subsection{Evaluation for Denoising Network}
\label{sec:eva_network}
In this section, we conduct experiments to evaluate the video denoising performance of our designed RVDT. 
We compare the effects of RVDT against current SOTA video denoising baselines on different datasets for a comprehensive evaluation, including PVDD, CRVD, and DAVIS datasets.

\begin{table}[t]
    \centering
    \huge
\caption{Quantitative comparison in the sRGB domain on both DAVIS and CRVD dataset.}
\label{comparison-rvdt-davis-crvd}
\resizebox{1.0\linewidth}{!}{
    \begin{tabular}{l|p{7.5cm}<{\centering}|p{7.5cm}<{\centering}}
        \toprule[1pt]
        & CRVD & DAVIS  \\
        \hline
        VNLnet & 22.62/0.873 & 28.40/0.907 \\
        FastDVDnet & 15.77/0.717 & 32.13/0.919  \\
        RViDeNet & 37.98/0.957 & 37.40/0.957 \\
        EDVR & 38.48/0.957 & 32.78/0.866 \\
        RVDT & \textbf{39.01/0.964} & \textbf{37.77/0.962} \\
        \bottomrule[1pt]
\end{tabular}}
\end{table}

\subsubsection{Evaluation Strategy}
We choose current SOTA video denoising frameworks as the baselines for comparison, as introduced in Sec.~\ref{sec:denosing_networks}.
First, we train our RVDT and all baselines on PVDD, and evaluate their performances on Dynamic20 and Static15 that have the corresponding ground truths. Then, RVDT and all baselines are trained on CRVD and DAVIS, and are evaluated on the corresponding testing datasets, respectively.

\begin{figure}[t]
    \centering
    \includegraphics[width=1.0\linewidth]{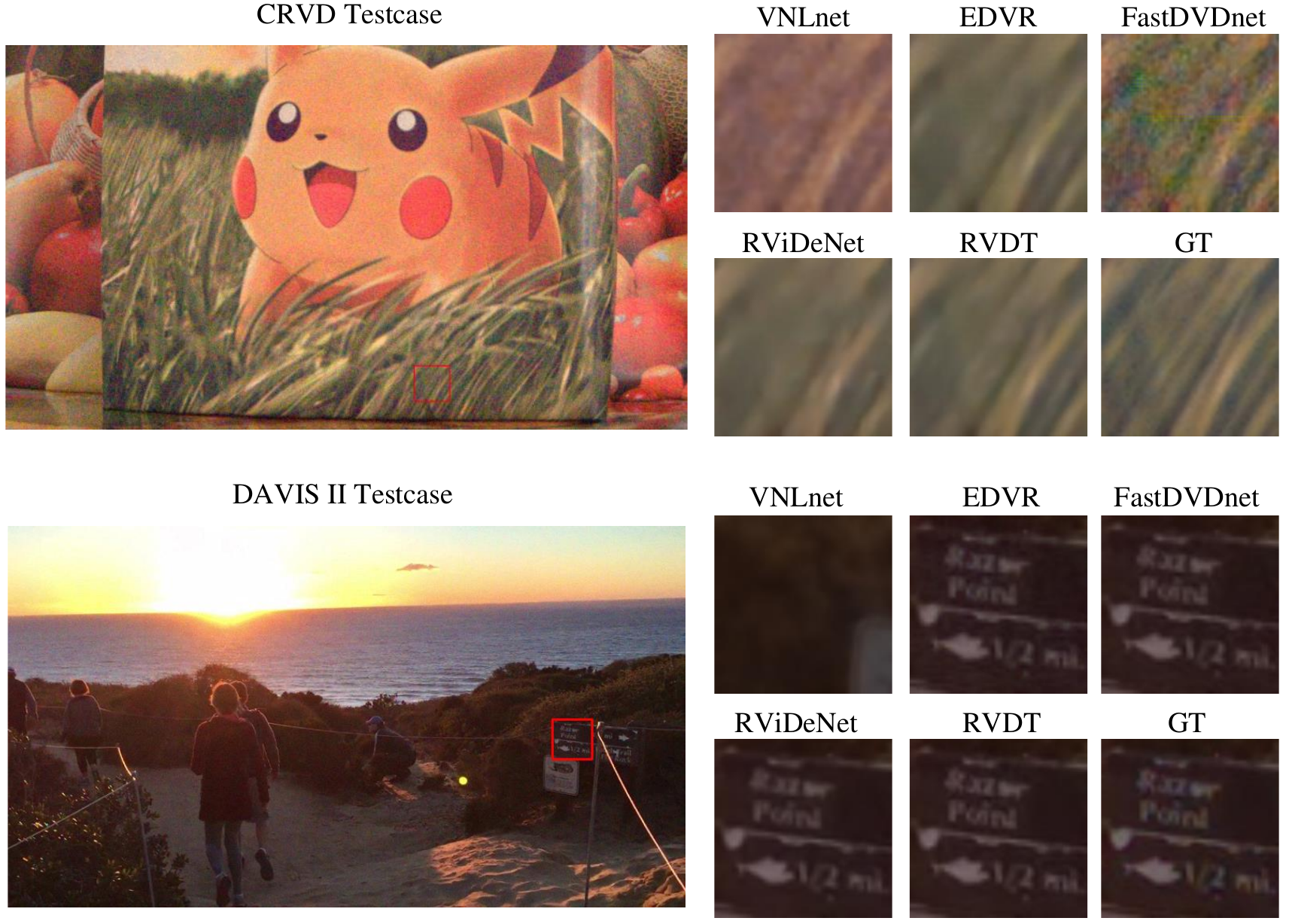}
    \caption{
The qualitative comparison for different networks (blind version) that are trained on CRVD and DAVIS datasets (sRGB domain). Our framework can achieve the SOTA performance on these two benchmarks.
    }
    \label{fig:network-crvd-davis}
\end{figure}

\subsubsection{Quantitative Results}
The results of the evaluation on the PVDD are shown in Table~\ref{comparison-rvdt-pvdd}, where all networks are trained with both blind and non-blind versions.
Trained on the same dataset, our designed RVDT obviously outperforms the other video-denoising approaches on both PSNR and SSIM, demonstrating the generalization ability of RVDT. 
For example, even under the challenging blind setting, RVDT achieves more than 1dB gain on the sRGB domain of Dynamic20 in terms of the PSNR, and more than 3\% advantage in terms of the SSIM.
The effectiveness of RVDT comes from the designed bi-directional spatial-temporal propagation mechanism achieved with the improved transformer blocks.

Moreover, the superiority of RVDT can also be proved when the comparison is conducted on the CRVD and DAVIS datasets.
As shown in Table~\ref{comparison-rvdt-davis-crvd}, our proposed RVDT can also achieve the SOTA performance, when trained on the training set of CRVD/DAVIS and evaluated on the corresponding testing set. The experiments are conducted in the sRGB domain and blind network version, which is the most challenging situation in practice. Moreover, the performance of RVDT is stable and is not dependent on the training data, and can be generalized to various types of testing data, resulting in satisfactory effects.

Furthermore, the superior performance of RVDT is not achieved by the large-scale model capability that is not practical. We compare the parameter number for RVDT and current efficient video denoising framework, e.g., FastDVDnet.
The parameter number of FastDVDnet is 8.239M, and ours is 2.487M, demonstrating the that RVDT owns the advantages in terms of the model efficiency.
With both SOTA denoising performance and inexpensive model parameter, RVDT is demonstrated as an effective and practical video denoising framework.

\subsubsection{Qualitative Results}
In Fig.~\ref{fig:network-srgb-blind}, we provide the visual comparison among different networks trained on the sRGB domain of PVDD. These networks are trained with blind versions. We can see that our RVDT generates the most pleasing and noise-free results with sharp and realistic details.
Moreover, Fig.~\ref{fig:network-srgb-noblind} shows the qualitative comparison among varying non-blind models. We can again see the advantage of our designed RVDT compared with current SOTA video denoising approaches.
The same visual superiority can also be observed in the RAW domain's comparison, as shown in Fig.~\ref{fig:network-raw-blind} (RAW images are processed with demisaicing for visualization).
Furthermore, we provide the qualitative comparison for networks trained on CRVD and DAVIS in Fig.~\ref{fig:network-crvd-davis}, proving the visual advantages of RVDT in noise removal and details preservation again.
The SOTA performance of RVDT on CRVD and DAVIS also demonstrates that the effectiveness of RVDT is not dependent on the training data.
All these visual samples demonstrate that RVDT is a practical and effective video denoising framework.

\begin{table}[t]
    \centering
    \huge
\caption{The results of ablation study.}
\label{comparison-abla}
\resizebox{1.0\linewidth}{!}{
    \begin{tabular}{l|p{7.5cm}<{\centering}|p{7.5cm}<{\centering}}
        \toprule[1pt]
        & Dynamic20  & Static15  \\
        \hline
        RVDT w/o CSA-MLP&34.76/0.926 &35.26/0.918 \\
        RVDT w/o Merge.& 35.28/0.936 & 35.58/0.917 \\
        RVDT w/o Trans.& 34.66/0.925 & 35.30/0.924 \\
        RVDT & \textbf{35.71/0.941}& \textbf{35.76/0.924 } \\
        \bottomrule[1pt]
\end{tabular}}
\end{table}

\subsubsection{Ablation Study}
We conduct a series of ablation studies to demonstrate the effectiveness of our proposed RVDT, by removing
different components from our framework individually.
We consider the following ablation settings.
\begin{enumerate}
    \item[a] RVDT w/o CSA-MLP: replace our designed CSA-MLP with the conventional MLP.
    \item[b] RVDT w/o Merge.: remove the ``Merging Layer" from the bi-directional temporal transformer block $\mathcal{T}_T$ and replace it with ``Transmission Layer".
    \item[c] RVDT w/o Trans.: remove the ``Transmission Layer" from the bi-directional temporal transformer block $\mathcal{T}_T$ and replace it with ``Merging Layer".
\end{enumerate}
The ablation studies are conducted on the sRGB domain and blind network version. The results are shown in Table~\ref{comparison-abla}.
Compared with all ablation settings, our full setting yields higher PSNR and SSIM. 
The results show the effects of the proposed CSA-MLP by comparing ``RVDT w/o CSA-MLP" and ``RVDT".
Moreover, the comparison among ``RVDT w/o Merge.", ``RVDT w/o Trans.", and ``RVDT" prove the superiority of simultaneously using the transmission layer and merging Layer to implement the temporal block rather than utilizing one of them only.

\section{Conclusion}
In this work, we have introduced to the community a large-scale dataset for video denoising in both RAW and sRGB domains, named PVDD. The significant contribution of this dataset is that it provides training videos with rich dynamic motion as well as realistic noise. 
Significantly, PVDD provides a practical benchmark for different video denoising networks in both sRGB and RAW domains, promoting the training of effective video denoising models.

Moreover, we propose a novel video denoising framework named RVDT with novelly designed bi-directional transformer blocks to achieve effective spatial-temporal propagation. Extensive comparisons are conducted on PVDD, CRVD, and DAVIS, demonstrating the superiority of RVDT over baselines.

\noindent\textbf{Future Work:}
Our dataset lacks semantic information. The semantic information, if available, could futher improve the video denoising effects. E.g., we can utilize the semantic mask to distinguish different areas in real-world videos and apply different degrees of denoising operations for these regions.
Therefore, in the future, we would enrich PVDD by adding more semantic information, e.g., providing semantic masks for different categories in video keyframes.

\bibliographystyle{abbrv}
\bibliography{egbib}

\begin{IEEEbiography}
        [{\includegraphics[height=1.25in,clip,keepaspectratio]{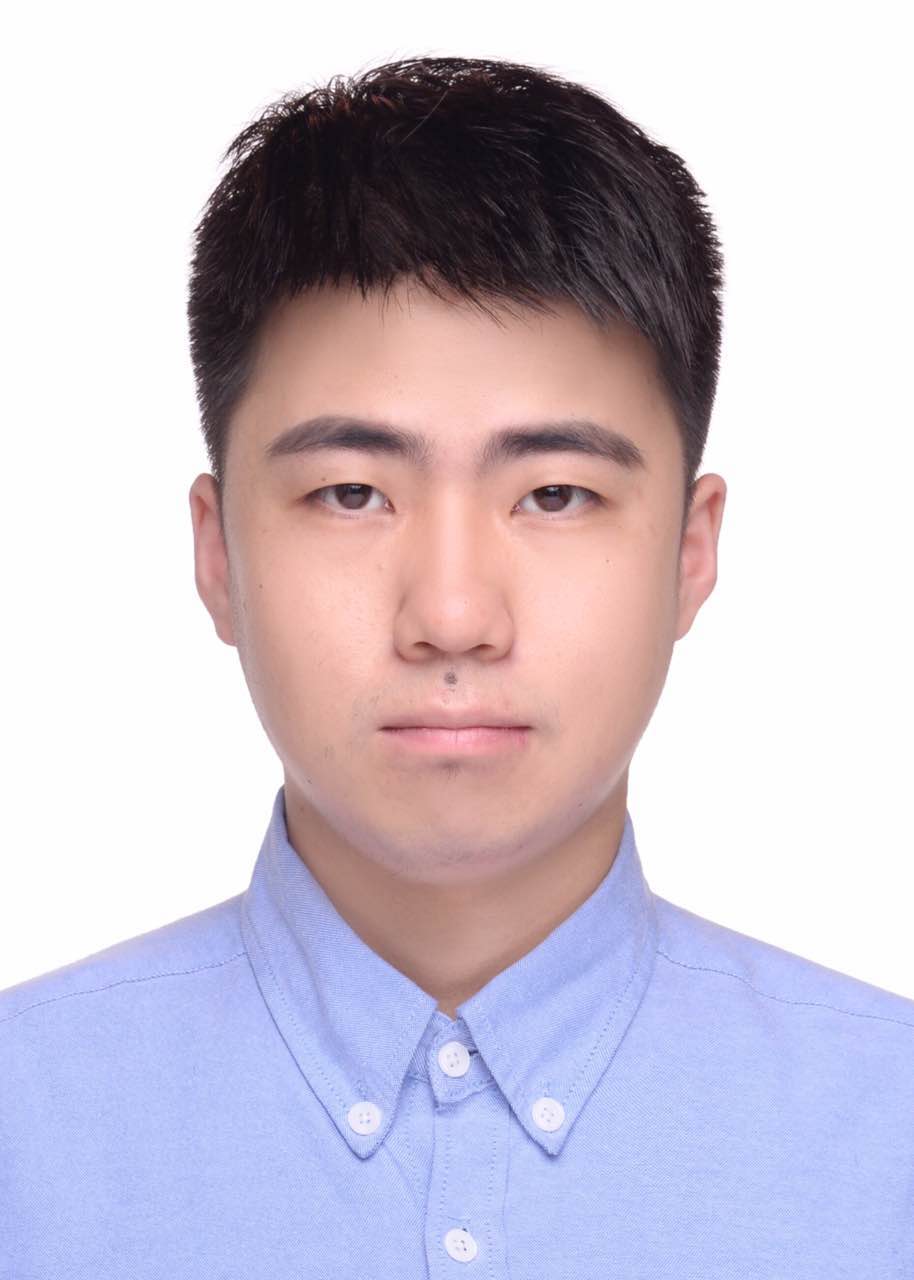}}]
        {Xiaogang Xu}
        is currently a fourth-year PhD student in the Chinese University of Hong Kong. He received his bachelor degree from Zhejiang University.
        He obtained the Hong Kong PhD Fellowship in 2018. He serves as a reviewer for CVPR, ICCV, ECCV, AAAI, ICLR, NIPS, IJCV. His research interest includes deep learning, generative adversarial networks, adversarial attack and defense, etc.
    \end{IEEEbiography}

\begin{IEEEbiography}
        [{\includegraphics[height=1.25in,clip,keepaspectratio]{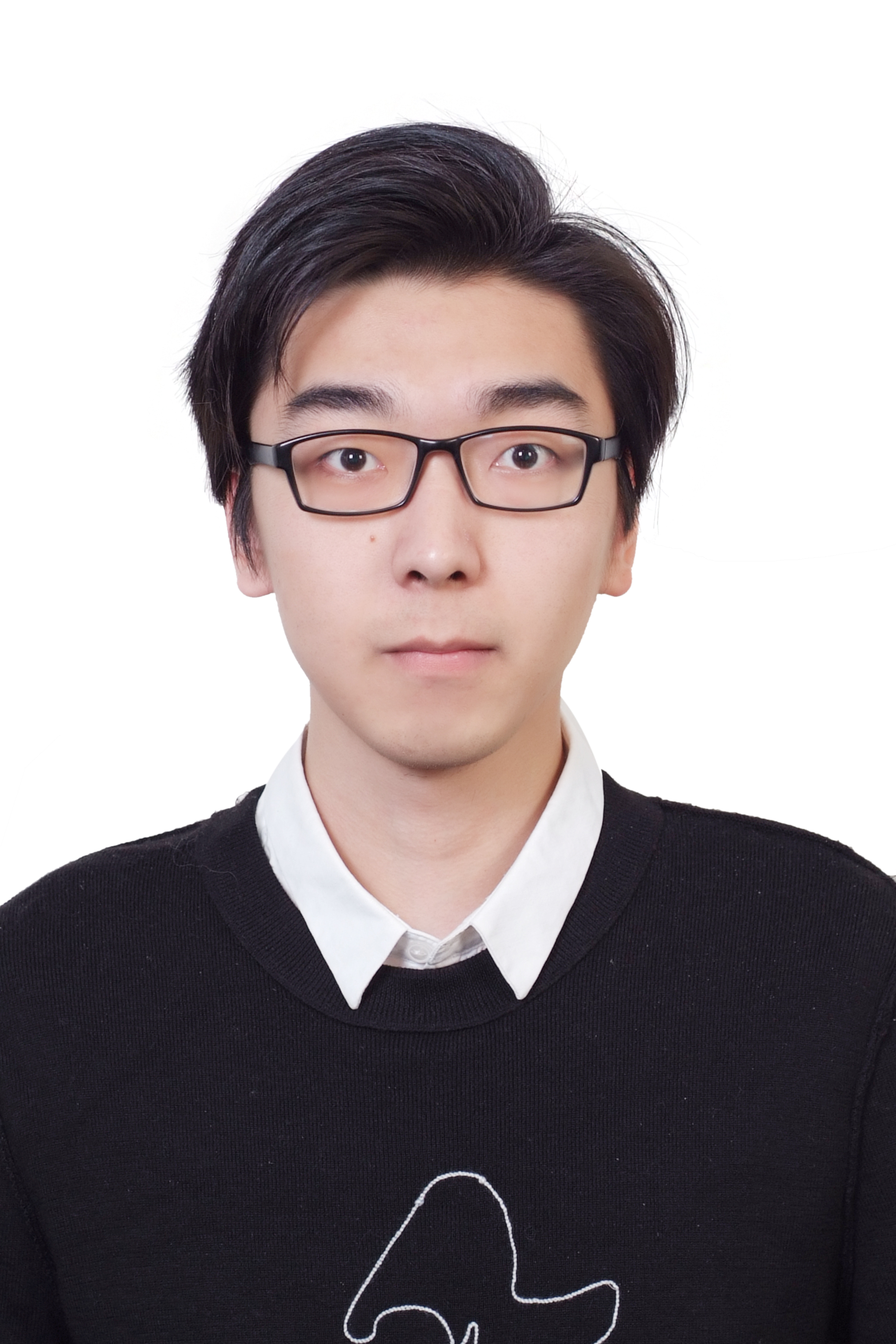}}]
        {Yitong Yu}
        is currently a computer vision researcher in the Mobile Intelligence Group of SenseTime. Before that, he received his bachelor degree from Central University of Finance and Economics. His research interest includes deep learning, image processing, knowledge distillation, etc.
    \end{IEEEbiography}


\begin{IEEEbiography}[{\includegraphics[width=1in,height=1.25in,clip,keepaspectratio]{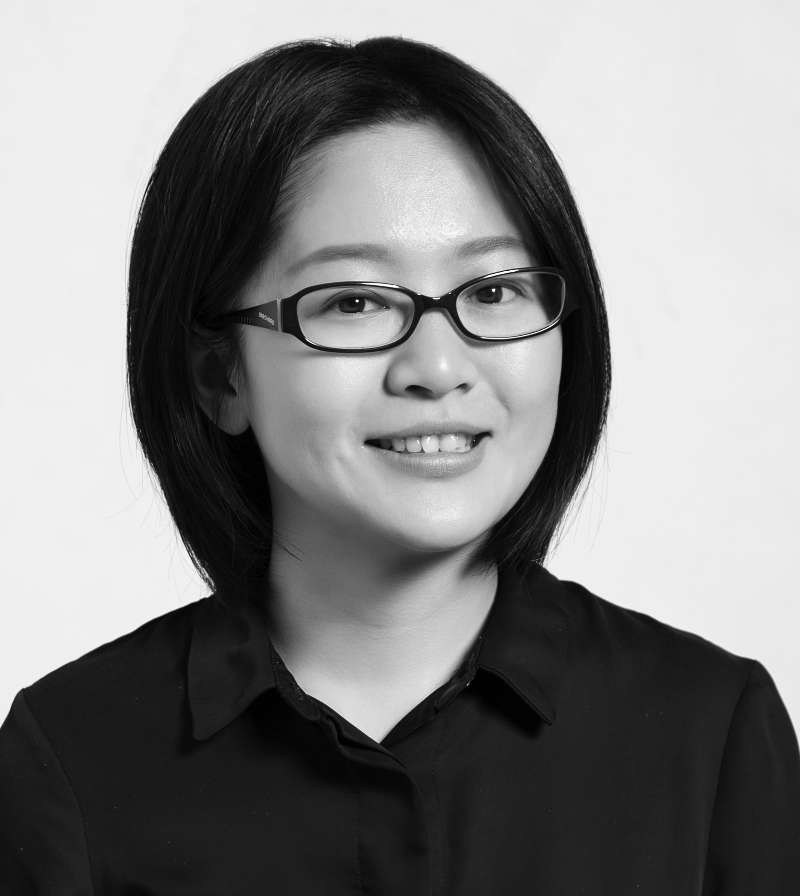}}]{Nianjuan Jiang} (M'12) received her B.S. and Ph.D degree in electrical and computer engineering from National University of Singapore, Singapore, in 2007 and 2012 respectively. From 2012 to 2016, she was with the Advanced Digital Sciences Center (ADSC), Singapore, which is a joint research center between the University of Illinois at Urbana-Champaign (UIUC), USA, and the Agency for Science, Technology and Research (A*STAR), Singapore. From 2017 to 2020, she was with Shenzhen Cloudream Technology Co., Ltd. as the R$\&$D director, where she was co-leading researchers and engineers to work on cutting-edge problems broadly in computer vision, computer graphics, and machine learning centered on perceiving, reconstructing and understanding humans as well as related innovative products. Since 2020, she is with SmartMore Co., Ltd., as a R$\&$D director. Her research interests include computer vision, computer graphics, 3D vision and computational photography. 
\end{IEEEbiography}

\begin{IEEEbiography}[{\includegraphics[height=1.26in,clip,keepaspectratio]{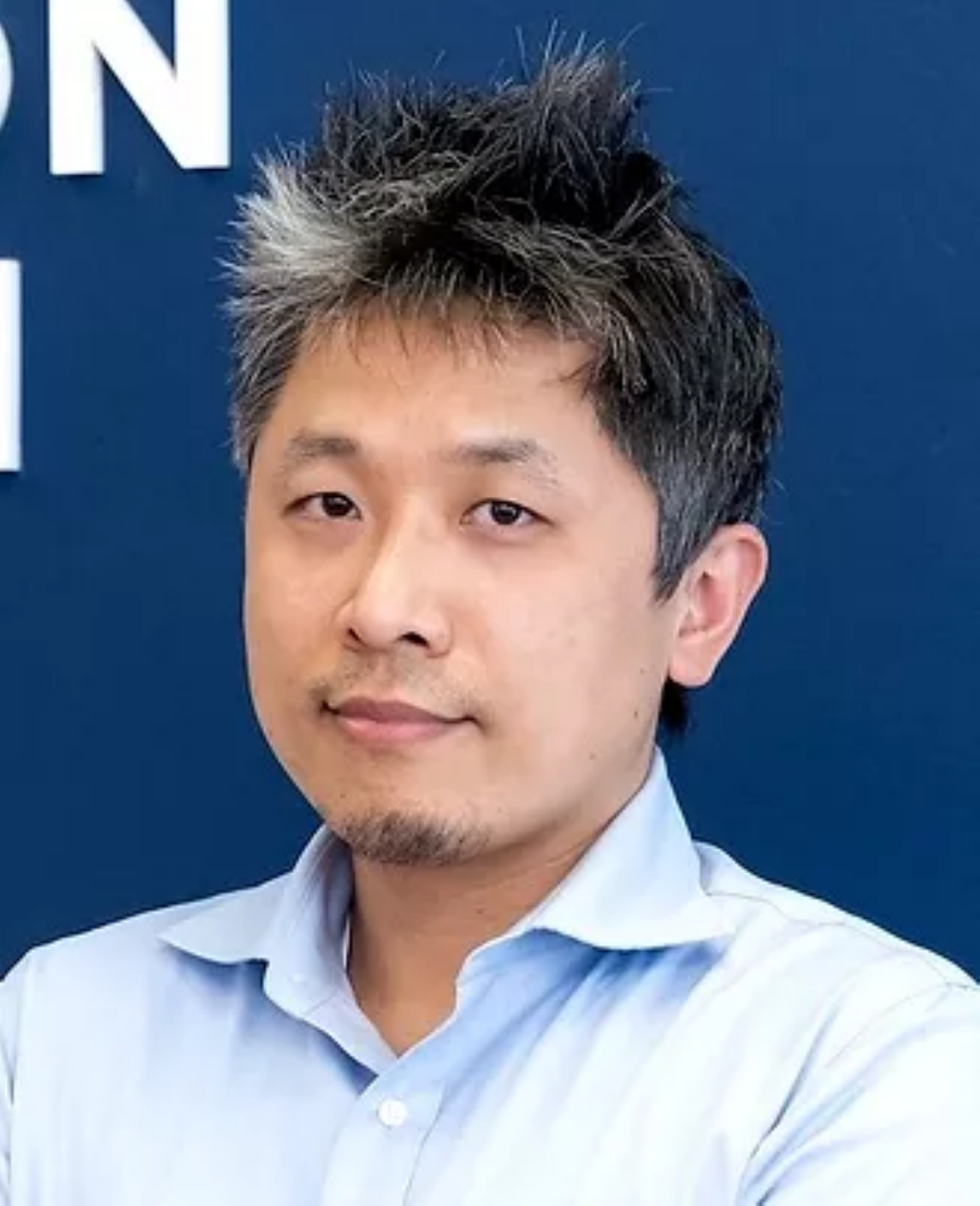}}]
    {Bei Yu}
    (M'15-SM'22)
    received the Ph.D.~degree from The University of Texas at Austin in 2014.
    He is currently an Associate Professor in the Department of Computer Science and Engineering, The Chinese University of Hong Kong.
    He has served as TPC Chair of ACM/IEEE Workshop on Machine Learning for CAD, and in many journal editorial boards and conference committees.
    He is Editor of IEEE TCCPS Newsletter.
    He received nine Best Paper Awards from DATE 2022, ICCAD 2021 \& 2013, ASPDAC 2021 \& 2012, ICTAI 2019, Integration, the VLSI Journal in 2018,
    ISPD 2017, SPIE Advanced Lithography Conference 2016, and seven ICCAD/ISPD contest awards.
\end{IEEEbiography}

\begin{IEEEbiography}[{\includegraphics[width=1in,height=1.25in,clip,keepaspectratio]{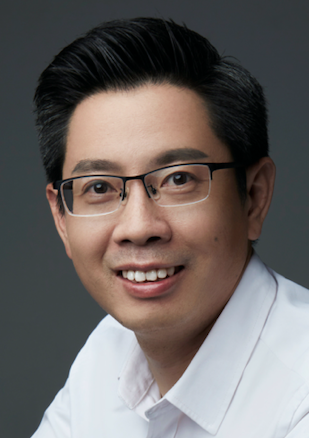}}]{Jiangbo Lu} (Senior Member, IEEE) received the Ph.D. degree from Katholieke Universiteit Leuven, Leuven, Belgium, in 2009. From 2009 to 2016, he was a Senior Research Scientist with the Advanced Digital Sciences Center (ADSC), a Singapore-based research center of University of Illinois at Urbana–Champaign (UIUC). From 2017 to 2020, he was with Shenzhen Cloudream Technology Co., Ltd., as the Chief Technology Officer (CTO) and as a Chief Scientist in 2019. Since 2020, he has been with SmartMore Co., Ltd., as the Co-founder and a CTO. Since 2021, he has also been an Adjunct Professor of the South China University of Technology. His research interests include computer vision, 3D vision, and image processing. 
\end{IEEEbiography}

    \begin{IEEEbiography}
        [{\includegraphics[height=1.25in,clip,keepaspectratio]{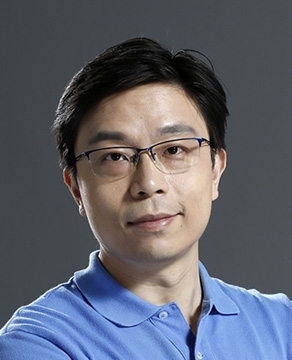}}]
        {Jiaya Jia}
        received the PhD degree in Computer Science from Hong Kong University of Science and Technology in 2004 and is currently a full professor in Department of Computer Science and Engineering at the Chinese University of Hong Kong (CUHK). He was a visiting scholar at Microsoft Research Asia from March 2004 to August 2005 and conducted collaborative research at Adobe Systems in 2007. He is an Associate Editor-in-Chief of IEEE Transactions on Pattern Analysis and Machine Intelligence (TPAMI) and  is also in the editorial board of International Journal of Computer Vision (IJCV). He continuously served as area chairs for ICCV, CVPR, AAAI, ECCV, and several other conferences for organization. He was on program committees of major conferences in graphics and computational imaging, including ICCP, SIGGRAPH, and SIGGRAPH Asia. He received the Young Researcher Award 2008 and Research Excellence Award 2009 from CUHK. He is a Fellow of the IEEE.
    \end{IEEEbiography}

\end{document}